\documentclass{article}

\PassOptionsToPackage{numbers}{natbib}
\usepackage{graphicx}
\usepackage{amsmath}
\usepackage{amssymb}
\usepackage{booktabs}

\usepackage{amsmath,amsfonts,bm}

\def\eqref#1{equation~\ref{#1}}

\def\1{\bm{1}}

\DeclareMathAlphabet{\mathsfit}{\encodingdefault}{\sfdefault}{m}{sl}
\SetMathAlphabet{\mathsfit}{bold}{\encodingdefault}{\sfdefault}{bx}{n}

\usepackage{url}

\usepackage{graphicx, color}
\usepackage{tikz}
\usepackage{comment}
\usepackage{amsmath,amssymb} %
\usepackage{color}
\usepackage{ragged2e}
\usepackage{xcolor}         %
\usepackage{graphicx}
\usepackage{multirow}

\usepackage{subcaption}

\usepackage{caption}
\usepackage{float}

\usepackage{url}            %
\usepackage{booktabs}       %
\usepackage{amsmath}
\usepackage{amsfonts}       %

\usepackage{nicefrac}       %
\usepackage{microtype}      %
\usepackage{nicefrac}       %
\usepackage{wrapfig}

\usepackage{multirow}
\usepackage{makecell}
\usepackage{threeparttable}
\usepackage{sidecap}
\usepackage{caption}

\usepackage{colortbl} %
\definecolor{mColor1}{rgb}{0.9,0.9,0.9}
\definecolor{mColor2}{rgb}{0.95,0.95,0.95}
\definecolor{non-photoblue}{rgb}{0.64, 0.87, 0.93}
\definecolor{lightblue}{rgb}{0.81, 0.94, 1.0}
\definecolor{DSBR_purple}{rgb}{0.58, 0.40, 0.74}
\usepackage{tikz}
\usepackage{pifont}

\definecolor{mColor1}{rgb}{0.9,0.9,0.9}
\definecolor{mColor2}{rgb}{0.95,0.95,0.95}
\definecolor{non-photoblue}{rgb}{0.64, 0.87, 0.93}
\definecolor{lightblue}{rgb}{0.81, 0.94, 1.0}
\definecolor{lightorange}{rgb}{0.965, 0.835, 0.71}

\definecolor{mygreen}{rgb}{0.000, 0.392, 0.000}

\newif\ifshowcomments
\showcommentstrue   %

\ifshowcomments
  
  \newcommand{\js}[1]{\textcolor{blue}{[\textbf{Julia}: #1]}}
  \newcommand{\dan}[1]{\textcolor{purple}{[\textbf{Daniel}: #1]}}
  \newcommand{\sam}[1]{\textcolor{orange}{[\textbf{Sameer}: #1]}}
  \newcommand{\tim}[1]{\textcolor{blue}{[\textbf{Tim}: #1]}}
\else
  
  \newcommand{\js}[1]{}
  \newcommand{\dan}[1]{}
  \newcommand{\sam}[1]{}
  \newcommand{\tim}[1]{}
\fi

\usepackage{titletoc}

\usepackage[numbers,sort&compress]{natbib}
\usepackage{wrapfig}
\usepackage{caption}
\usepackage{algorithmic} %

\usepackage{wrapfig}
\usepackage{orcidlink}
\definecolor{mypurple}{rgb}{0.502, 0.000, 0.502}

\newcommand{\bluetext}[1]{\textcolor{blue}{#1}}
\newcommand{\browntext}[1]{\textcolor{brown}{#1}}
\newcommand{\purpletext}[1]{\textcolor{magenta}{#1}}

\renewcommand{\bluetext}[1]{#1}
\renewcommand{\browntext}[1]{#1}
\renewcommand{\purpletext}[1]{#1}

\usepackage{algorithm}
\usepackage{pifont}

\usepackage[capitalize]{cleveref}
\crefname{section}{Sec.}{Secs.}
\Crefname{section}{Section}{Sections}
\Crefname{table}{Table}{Tables}
\crefname{table}{Tab.}{Tabs.}

\usepackage{booktabs} %
\usepackage{multirow} %
\usepackage{graphicx} %

\usepackage{soul}
\usepackage{xcolor}

\sethlcolor{yellow} %

\definecolor{cgreen}{rgb}{0.2,0.6,1}

\definecolor{darkgreen}{RGB}{0,100,0}

\usepackage{etoc}

\usepackage{amsmath,amssymb}
\usepackage{bm} %

\usepackage{colortbl} %
\definecolor{mColor1}{rgb}{0.9,0.9,0.9}
\definecolor{mColor2}{rgb}{0.95,0.95,0.95}
\definecolor{non-photoblue}{rgb}{0.64, 0.87, 0.93}
\definecolor{lightblue}{rgb}{0.81, 0.94, 1.0}
\definecolor{lightorange}{rgb}{0.965, 0.835, 0.71}

\usepackage{booktabs}
\usepackage{multirow}
\usepackage[table]{xcolor}
\usepackage{amssymb}

\newcommand{\pos}[1]{\textcolor{green!60!black}{\(\textstyle\blacktriangle\)}\,#1}
\newcommand{\nega}[1]{\textcolor{red}{\(\textstyle\blacktriangledown\)}\,#1}
\newcommand{\neu}[1]{\textcolor{gray}{\(\textstyle\blacktriangle\)}\,#1}

\usepackage[table,dvipsnames]{xcolor}
\usepackage{booktabs}
\usepackage{multirow}
\usepackage{xfp}

\newcommand{\MaxDelta}{6}

\newcommand{\ScoreDelta}[2]{%
  \begingroup
  \edef\shade{\fpeval{round(min(85,10 + 75*abs(#2)/\MaxDelta),0)}}%
  \ifdim #2 pt > 0.05pt
    \cellcolor{green!\shade!white} #1 {\scriptsize\textcolor{green!50!black}{(+#2)}}%
  \else\ifdim #2 pt < -0.05pt
    \cellcolor{red!\shade!white} #1 {\scriptsize\textcolor{red!70!black}{(#2)}}%
  \else
    \cellcolor{gray!18} #1 {\scriptsize\textcolor{black!65}{(+0.0)}}%
  \fi\fi
  \endgroup
}

\newcommand{\AvgDelta}[1]{%
  \begingroup
  \edef\shade{\fpeval{round(min(85,10 + 75*abs(#1)/\MaxDelta),0)}}%
  \ifdim #1 pt > 0.05pt
    \cellcolor{green!\shade!white}{\scriptsize\textcolor{green!50!black}{(+#1)}}%
  \else\ifdim #1 pt < -0.05pt
    \cellcolor{red!\shade!white}{\scriptsize\textcolor{red!70!black}{(#1)}}%
  \else
    \cellcolor{gray!18}{\scriptsize\textcolor{black!65}{(+0.0)}}%
  \fi\fi
  \endgroup
}

\usepackage[table]{xcolor}      %
\usepackage{pifont}             %

\newcommand{\statuscollapse}{%
  \multicolumn{2}{c}{\textbf{\textcolor{orange!75!black}{\ding{56}\ {Model Collapse}}}}}
\newcommand{\statusstable}{%
  \multicolumn{2}{c}{\textit{\textcolor{orange!55}{\ding{51}\ Stable}}}}

\usepackage{xcolor}
\usepackage{amssymb}

 \usepackage[main, final]{neurips_2026}

\usepackage[utf8]{inputenc} %
\usepackage[T1]{fontenc}    %
\usepackage{hyperref}       %
\usepackage{url}            %
\usepackage{booktabs}       %
\usepackage{amsfonts}       %
\usepackage{nicefrac}       %
\usepackage{microtype}      %
\usepackage{xcolor}         %

\usepackage{multirow}
\usepackage{float}
\DeclareMathOperator*{\argmax}{arg\,max}

\title{Entropy Minimization without Model Collapse: Mitigating Prediction Bias in Medical Imaging}

\author{
  Tim Nielen\textsuperscript{1}\footnotemark[1] ,
  Sameer Ambekar\textsuperscript{1,2,4,5}\footnotemark[1] ,
  Daniel M. Lang\textsuperscript{1,2}\textsuperscript{\#} ,
  \textbf{Julia A. Schnabel}\textsuperscript{1,2,3,4,5}\textsuperscript{\#} \\
  \textsuperscript{1}School of Computation, Information and Technology, Technical University of Munich, Germany \\
  \textsuperscript{2}Institute of Machine Learning in Biomedical Imaging, Helmholtz Munich, Germany \\
  \textsuperscript{3}School of Biomedical Engineering and Imaging Sciences, King's College London, UK \\
  \textsuperscript{4}Munich Center for Machine Learning (MCML) \\
  \textsuperscript{5}relAI -- Konrad Zuse School of Excellence in Reliable AI \\
}

\begin{document}

\maketitle
\renewcommand{\thefootnote}{\fnsymbol{footnote}}
\footnotetext[1]{Shared first-authorship \quad \textsuperscript{\#}Shared last-authorship}
\begin{abstract}
    Entropy minimization (EM) is the dominant objective for test-time adaptation, yet its failure mode, model collapse, remains poorly understood. In this work, we show that distribution shifts can cause feature clusters corresponding to distinct classes in the model’s representation space to merge, while the decision boundary remains fixed. This induces a systematic skew in the predicted class distribution, referred to as prediction bias. Prediction bias refers to a shift in the predicted class distribution, with some classes overrepresented and others suppressed. We show that entropy minimization amplifies this prediction bias by tightening the existing clusters, reinforcing the incorrect groupings until all predictions collapse to a trivial solution. Next, to demonstrate the significance of prediction bias and mitigate it, we further propose \textit{Distribution Shift Bias Reduction} (DSBR), a bias-correcting objective that specifically targets this failure mode by equalizing the contribution of each predicted class to the unsupervised entropy minimization loss.
To study this failure mode, we design suitable adaptation settings using four medical‑imaging datasets and additionally evaluate on ImageNet‑C. We find that DSBR consistently stabilizes test‑time adaptation, prevents model collapse, and matches or outperforms the state‑of‑the‑art methods. Moreover, DSBR operates solely at test-time. 
Our code is available at \url{https://entropy-minimization-dsbr.github.io/}.

\end{abstract}
\section{Introduction}

Machine learning models face a fundamental challenge during real-world deployment: a model trained on source data must adapt to unseen data at inference time, without access to the training data or labels from the unseen deployment domain~\cite{xiao2024beyond,wang2018deep,chen2022contrastive}. This challenge arises through distribution shifts caused by, e.g., changes in sensor properties~\cite{yoon2024domain_mi}, environmental conditions~\cite{hoffman2018cycada}, and acquisition protocols~\cite{yoon2024domain_mi} that alter the input data distribution while leaving the semantic task itself unchanged. 
Such shifts can lead to substantial drops in model performance when the learned representation is no longer well aligned with the target data~\cite{pandey2020unsupervised,gulrajani2021in,iwasawa2021test}. 
Medical imaging is a particularly important example of this setting~\cite{van2021deep}. 
A pathology image acquired with a different scanner or stain remains semantically the same, yet becomes statistically misaligned with the source domain \citep{karani2021test, stacke2020measuring}. 
In these scenarios, the model's confidence and prediction reliability can deteriorate significantly~\cite{koh2021wilds, stacke2020measuring,balendran2025scoping}. 

\begin{figure}[t]
    \centering
    \includegraphics[width=\linewidth]{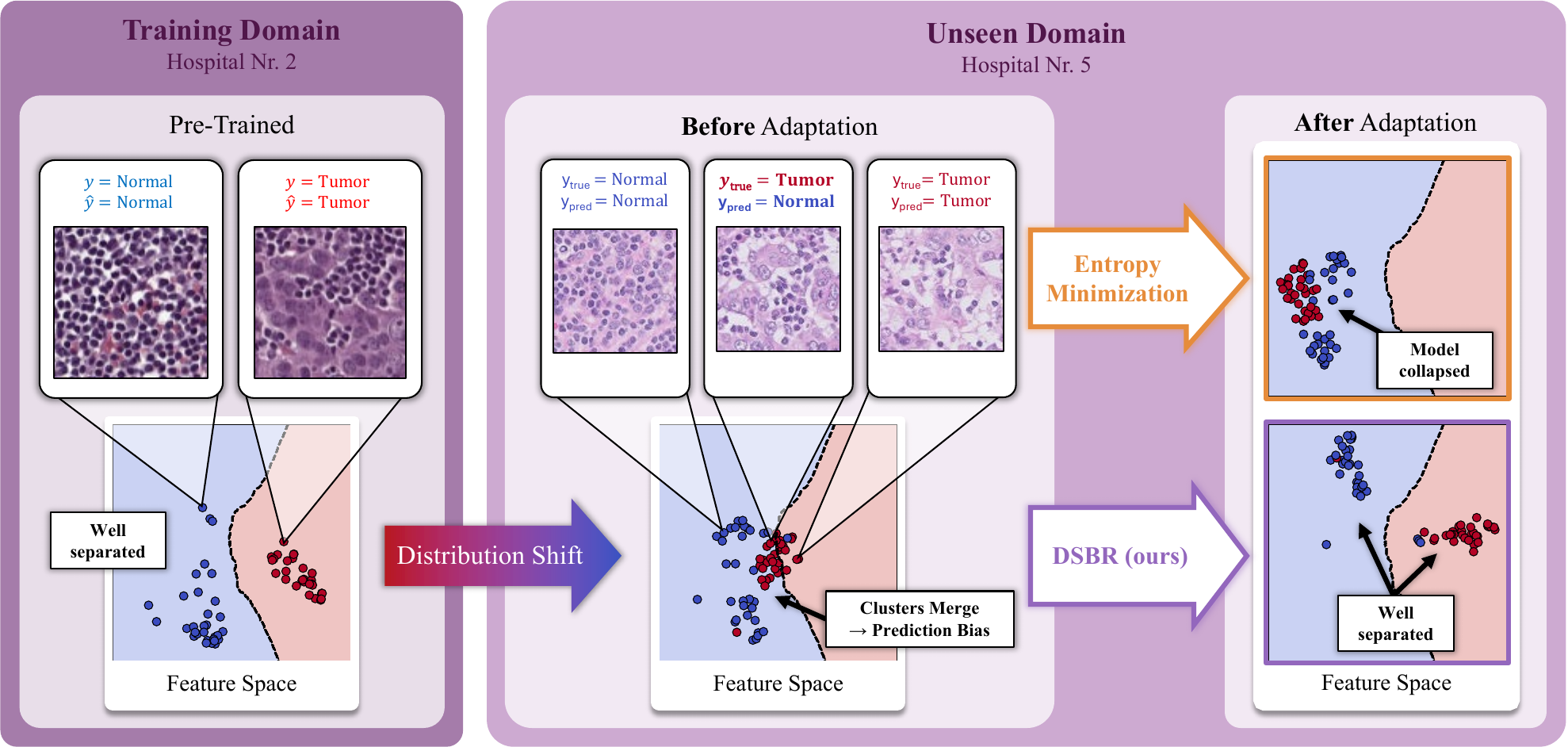}
    \includegraphics[width=\linewidth]{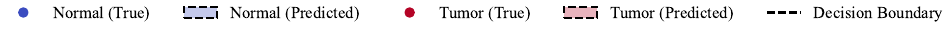}
    \caption{\textbf{Feature space clusters merge under distribution shift,
inducing severe prediction bias, which is prone to failure for Entropy Minimization (EM).} We show the 2-dimensional t-SNE embeddings from WILDSCamelyon~\citep{koh2021wilds} dataset featurized by ResNet50-GN. The background colors indicate the area of the feature space the model classifies as each class which we obtain by fitting a $k$-nearest neighbors classifier on the 2D embeddings and corresponding model predictions. Entropy minimization \cite{wang2021tent,lee2024entropy} is prone to model collapse. In contrast, \textbf{DSBR (ours)} mitigates model collapse and prevents cluster merging across multiple iterations. }
    
    \label{fig:domain_shift_bias}
\end{figure}

In natural vision, Entropy Minimization (EM) has emerged as the dominant objective for Test-Time Adaptation (TTA) methods designed to handle distribution shifts during deployment~\cite {press2024entropy,wang2021tent,liu2021ttt++,niu2023towards}. A seminal work on EM, {Tent} \citep{wang2021tent} showed that minimizing the Shannon entropy~\cite{shannon1948mathematical} of the model's predictions over the affine parameters of batch normalization layers reliably reduces prediction error under mild distribution shift. Since then, EM-based unsupervised loss has been widely used as an unsupervised objective for many TTA methods \citep{niu2022efficient, niu2023towards, zhang2025come,zhang2021memo}.  
Subsequent work has focused on making EM more robust in {wild} scenarios, including limited batch sizes and mixed distribution shifts~\citep{niu2023towards}, through filtering unreliable samples \citep{niu2023towards, lee2024entropy}, modified loss
functions \citep{zhang2025come}, sharpness-aware optimization~\citep{niu2023towards}, and preventing model collapse~\cite{niu2023towards}. With model collapse depicting a critical failure mode in which the model outputs only one or a few classes, irrespective of the input
~\cite{dohmatob2025strong,dohmatob2024model}. 

The progress of TTA methods for natural‑image tasks has been largely driven by benchmarks that introduce synthetic distribution shifts in the form of severe corruptions or do not fully capture the real-world shifts~\cite{hendrycks2019benchmarking,peng2017visda,yu2023benchmarking}. 
In contrast, the shifts encountered in medical imaging are typically much subtler~\cite{kilim2022physical}. They often preserve the underlying pathology while being driven by clinically relevant factors such as changes in patient demographics, scanner hardware, staining protocols, or acquisition parameters.
These variations modify the input statistics in ways that are difficult to detect visually~\cite{yoon2024domain_mi}, thereby posing a challenging problem for TTA methods.
Moreover, as stated above, model collapse poses a critical risk because it produces single-dominant class outputs regardless of the input~\cite{dohmatob2025strong,dohmatob2024model}. In clinical settings, such failures may go unnoticed in practice, allowing compromised models to influence decisions in ways that can lead to serious, potentially life-threatening effects~\cite{bernhardt2022failure,evans2024understanding,wiens2019no,lekadir2025future}.

Therefore, a rethinking of entropy minimization for medical imaging is needed.
In this setting, deployment shifts are not only unknown in advance but also subtle and heterogeneous across acquisition protocols, scanners, and patient populations~\cite{yoon2024domain_mi,kilim2022physical}.
 Consequently, heuristics that perform well on synthetic corruption benchmarks do not transfer reliably to the medical domain~\citep{evans2024understanding,kilim2022physical,roschewitz2023automatic}.
While ~\citet{press2024entropy} show that entropy minimization can succeed or fail under distribution shifts, the mechanisms driving model collapse remain unclear. In particular, \purpletext{it remains unclear {from a theoretical perspective} why the same EM objective improves adaptation under some shifts yet leads to catastrophic collapse under others.}

As illustrated in \cref{fig:domain_shift_bias}, distribution shift distorts the feature space geometry. As class clusters drift and partially merge while the classifier boundary remains fixed, prediction bias amplifies, with some classes absorbing ambiguous samples and others becoming suppressed. To understand and improve entropy minimization, we make the following contributions:

\begin{itemize}
    \item \textbf{Prediction bias as the root cause of model collapse, amplified by entropy minimization.}
    We identify prediction bias, a skew in the predicted class distribution induced by distribution shift, as the central failure mode of entropy minimization. We show that the shift distorts feature geometry, causing class clusters to drift and partially merge while the classifier boundary remains fixed. Entropy minimization amplifies this bias by reinforcing already dominant clusters, ultimately leading to model collapse.

    \item \textbf{Mitigating model collapse with Distribution Shift Bias Reduction (DSBR).}
    We propose DSBR, an importance reweighting method that equalizes the contributions of each predicted class to the entropy loss. By preventing primarily predicted classes from dominating the optimization signal, DSBR mitigates prediction bias and promotes a more stable separation of feature clusters during adaptation, as shown in \cref{fig:domain_shift_bias}.

    \item \textbf{We design challenging entropy-minimization scenarios for medical imaging.}
    We evaluate state-of-the-art baselines and DSBR on challenging entropy-minimization settings, including clinically grounded scenarios and clinical shifts in medical imaging. In addition, we demonstrate the effectiveness of DSBR in preventing model collapse. 
\end{itemize}

\section{Background and Related Work}\label{sec:background_related_work}

\textbf{Notations.} We focus on the task of image classification. Let $\mathcal{X}$ denote the input space, where random samples $X$ are distributed according to a probability distribution $p_X(x)$ for $x \in \mathcal{X}$, and let $\mathcal{K} := \{1,\dots,K\}$ denote the set of classes. We consider a pre-trained model $f_{\theta} : \mathcal{X} \to \Delta^{K-1}$ with parameters $\theta$, and write it as the composition $f_{\theta}(x) = h(\phi(x))$, where $\phi$ is a featurizer mapping inputs to a representation space and $h$ is a classifier acting on these representations. We define the per-sample entropy loss as $l(x) := H(f_{\theta}(x))$ and the predicted class as $\hat{y}(x) := \argmax f_{\theta}(x)$. Finally, $p_{\hat Y}(k)$ describes the predicted class distribution given $\hat Y = \hat y(X)$ and $k\in \mathcal{K}$.

\textbf{{The Entropy Minimization Enigma.}} In test-time adaptation, a model trained on source data, must adapt to
unlabeled test data at inference time, without access to source data or any labels
\citep{wang2021tent}. \purpletext{Unlike offline adaptation methods, including source-free domain adaptation~\cite{xiao2024beyond,liang2025comprehensive}, which typically assume access to the entire target set, test-time adaptation operates solely online on a stream of incoming test data.} Approaches include, but are not limited to, updating batch normalization statistics using an unsupervised loss \citep{schneider2020improving, nado2020evaluating},
leveraging surrogate tasks to learn test-time representations \citep{sun2020test}, utilizing pseudo-labels \citep{lee2013pseudo}, and minimizing the entropy of model predictions
\citep{wang2021tent}. Among these, entropy minimization (EM) has become the dominant choice for a wide range of methods~\citep{niu2022efficient, niu2023towards, wang2022continual, zhang2025come}, and is also the focus of this work. Specifically, it minimizes the Shannon entropy~\cite{shannon1948mathematical} of the softmax output, $H(\mathbf{p}) = -\sum_{i=1}^{K} p_i \log p_i$
where $\mathbf{p} = f_\theta(\mathbf{x}) \in \Delta^{K-1}$.

This equation assumes that lower-entropy predictions correlate with higher accuracy \citep{wang2021tent}.
{Tent} \citep{wang2021tent} uses EM to update the affine
parameters of batch normalization layers and achieves strong results on image corruption
benchmarks such as ImageNet-C and CIFAR-10/100-C. However, \citet{niu2023towards}
showed that {Tent} is prone to failure in more realistic \textit{wild} TTA scenarios. Overall, the inner mechanisms driving both the successes and
failures of EM remain poorly understood, a phenomenon \citet{press2024entropy} termed \textit{Entropy Enigma}. They showed that EM acts as a cluster refinement process, tightening feature representations within clusters of predicted classes, and identified a failure case where these clusters drift away from their original class centers. We show that this failure is even seeded before adaptation begins: distribution shifts can cause clusters of distinct classes to merge, leading EM to reinforce incorrect groupings.

\textbf{{Stabilizing Entropy Minimization.} }The standard approach to mitigate EM failure is to identify and discard unreliable
samples \citep{niu2022efficient, niu2023towards, lee2024entropy}. A common proxy for reliability is prediction entropy: samples above a
fixed entropy threshold are considered too uncertain and excluded from the backward pass
\citep{niu2022efficient, niu2023towards, lee2024entropy}. DeYO
\citep{lee2024entropy}  argues that entropy alone is an
unreliable signal and augments it with a shape-based score that measures how much
a structure-destroying augmentation changes the model's prediction.
Notably, such methods require dataset- and model-dependent hyperparameters to be fixed. However, high-entropy
samples are not always harmful as stated in \citet{press2024entropy}. Moreover, filtering is wasteful and becomes problematic in scenarios where data is scarce, which is often the case for medical imaging. Several methods further stabilize
EM by modifying the loss or the optimization.
EATA \citep{niu2022efficient} adds a Fisher information-based regularizer to prevent the model from forgetting its source-domain knowledge during adaptation.
SAR \citep{niu2023towards} applies sharpness-aware minimization
\citep{foret2021sharpness} to encourage convergence to flat minima, making
updates more robust to noisy gradients. ROID \citep{marsden2024universal}
maintains a running weight ensemble between the adapted and source model to
prevent parameter drift. \purpletext{However, ROID, and mean-teacher methods in general, do not address the root cause of model collapse, and instead solve it by incurring increased compute costs due to auxiliary teacher and augmentations.} Recently, COME \citep{zhang2025come} replaces the entropy loss
with one derived from a Dirichlet prior and down-weights the low-evidence predictions. \bluetext{In addition, energy-based works, including ~\citet{park2026rethinking}, augment the objective with additional loss terms for optimization.}  We deliberately avoid all of these strategies for our method, as our goal is to isolate and directly target prediction bias as the primary driver of EM failure.

\textbf{Importance Reweighting.} Importance reweighting is also used to address distribution shifts~\citep{shimodaira2000improving,huang2006correcting,sugiyama2007covariate,gretton2009covariate,lipton2018detecting}. Existing work addresses input shift by reweighting the pre-training loss with $p_X(x)/q_X(x)$~\citep{shimodaira2000improving,huang2006correcting,sugiyama2007covariate,gretton2009covariate}, but this requires access to source data and it's distribution $q_X$ and is thus not applicable in fully test-time adaptation. 
In our setting, input shift induces a bias in the predicted class distribution $p_{\hat Y}$. To target this, ROID~\citep{marsden2024universal} performs diversity weighting through cosine dissimilarity over the model's softmax outputs. These are known to be miscalibrated in modern network architectures~\cite{guo2017calibration}. We however, reweight the entropy loss proportional to $1/p_{\hat Y}$ directly.

\section{Prediction Bias Leads to Model Collapse and Is Amplified by Entropy Minimization}\label{sec:prediction_bias}

\subsection{Prediction Bias Is the Root Cause of Model Collapse}\label{sec:bias_root}

\begin{table}[t]
\centering \tiny
\setlength{\abovecaptionskip}{3pt}
\setlength{\belowcaptionskip}{0pt}

\begin{subtable}[t]{0.44\textwidth}
\caption{\textbf{WILDSCamelyon}: Hospital Nr. 2 $\rightarrow$ 4}
\label{tab:WILDSCamelyon_mild}
\centering
\resizebox{\linewidth}{!}{%
\begin{tabular}{lccc}
\toprule
\statusstable & \multicolumn{2}{c}{\textbf{Predicted}} \\
\cmidrule(lr){1-2}\cmidrule(lr){3-4}
\textbf{Class} & \textbf{True} & \textbf{Before EM} & \textbf{After EM} \\
\midrule
Normal & \cellcolor{orange!50} 49.8\% & \cellcolor{orange!63} 62.8\% & \cellcolor{orange!52} 52.4\% \\
Tumor  & \cellcolor{orange!50} 50.2\% & \cellcolor{orange!37} 37.2\% & \cellcolor{orange!48} 47.6\% \\
\bottomrule
\end{tabular}}
\end{subtable}\hfill
\begin{subtable}[t]{0.52\textwidth}
\caption{\textbf{ImageNet-C}: Gaussian Noise level 3}
\label{tab:ImageNetC_mild}
\centering
\resizebox{\linewidth}{!}{%
\begin{tabular}{lccc}
\toprule
\statusstable & \multicolumn{2}{c}{\textbf{Predicted}} \\
\cmidrule(lr){1-2}\cmidrule(lr){3-4}
\textbf{Class} & \textbf{True} & \textbf{Before EM} & \textbf{After EM} \\
\midrule
Christmas stocking  & \cellcolor{orange!0.1} 0.1\% & \cellcolor{orange!0.1} 0.2\% & \cellcolor{orange!0.1} 0.3\% (max) \\
999 other (median)  & \cellcolor{orange!0.1} 0.1\% & \cellcolor{orange!0.1} 0.1\% & \cellcolor{orange!0.1} 0.1\% \\
\bottomrule
\end{tabular}}
\end{subtable}

\vspace{1em}

\begin{subtable}[t]{0.44\textwidth}
\caption{\textbf{WILDSCamelyon}: Hospital Nr. 2 $\rightarrow$ 5}
\label{tab:WILDSCamelyon_severe}
\centering
\resizebox{\linewidth}{!}{%
\begin{tabular}{lccc}
\toprule
\statuscollapse & \multicolumn{2}{c}{\textbf{Predicted}} \\
\cmidrule(lr){1-2}\cmidrule(lr){3-4}
\textbf{Class} & \textbf{True} & \textbf{Before EM} & \textbf{After EM} \\
\midrule
Normal & \cellcolor{orange!50} 49.6\% & \cellcolor{orange!85} 85.0\% & \cellcolor{orange!100} \textbf{100.0\%} \\
Tumor  & \cellcolor{orange!50} 50.4\% & \cellcolor{orange!15} 15.0\% & \cellcolor{orange!0.0} \textbf{0.0\%} \\
\bottomrule
\end{tabular}}
\end{subtable}\hfill
\begin{subtable}[t]{0.52\textwidth}
\caption{\textbf{ImageNet-C}: Fog level 5}
\label{tab:ImageNetC_severe}
\centering
\resizebox{\linewidth}{!}{%
\begin{tabular}{lccc}
\toprule
\statuscollapse & \multicolumn{2}{c}{\textbf{Predicted}} \\
\cmidrule(lr){1-2}\cmidrule(lr){3-4}
\textbf{Class} & \textbf{True} & \textbf{Before EM} & \textbf{After EM} \\
\midrule
Geyser              & \cellcolor{orange!0.2} 0.2\% & \cellcolor{orange!26} 25.5\% & \cellcolor{orange!99} \textbf{99.3\%} \\
999 other (median)  & \cellcolor{orange!0.1} 0.1\% & \cellcolor{orange!0.1} 0.1\% & \cellcolor{orange!0.0} 0.0\% \\
\bottomrule
\end{tabular}}
\end{subtable}

\caption{\textbf{Prediction bias leads to Model collapse.}
Predicted class distributions before and after EM with ResNet50-GN pre-trained on
WILDSCamelyon Hospital~2 (a,c) and ImageNet (b,d), on 20\% holdouts of respective
target domains. The color shadings indicate the prediction bias; the greater the color difference between \emph{True} and \emph{Before EM}, the more likely it is that model collapse occurs. }
\label{tab:pred_bias}
\end{table}

Model collapse under entropy minimization is rooted in a skew of the predicted class prior on the target domain. We refer to this skew as \emph{prediction bias}, consistent with prior work~\citep{marsden2024universal,lee2024entropy,deng2026panda,wang2026partitionthenadapt}.
Our central claim, however, is different: under distribution shift, prediction bias is already present before adaptation, and during adaptation, entropy minimization subsequently amplifies it until collapse.

To explain the origin of this bias, consider a standard model of the form $f_\theta(x)=h(\phi(x))$, where $\phi$ is a featurizer and $h$ is a classifier. After source training, the learned representation typically organizes samples of different classes into compact and approximately separable regions in feature space. The classifier $h$ then induces a decision boundary
\[
\mathcal{H}:=\{z\in\phi(\mathcal{X}) \mid h(z)_i = h(z)_j,\; i\neq j\},
\]
which partitions the feature space into regions associated with different predicted classes. Under distribution shift, however, the feature distribution changes while the classifier remains fixed. Some class clusters remain stable, whereas others drift toward neighboring classes and partially cross the decision boundary. This produces a skew in the model's predictions: certain classes absorb ambiguous samples from others, becoming overrepresented, while others are suppressed. Crucially, this prediction bias is present before any adaptation takes place.

\begin{figure}[t]
     \centering

     \begin{subfigure}[b]{\textwidth}
        \small
         \begin{subfigure}[b]{0.192\textwidth}
             \centering
             Before Adaptation\\
             \includegraphics[width=\textwidth]{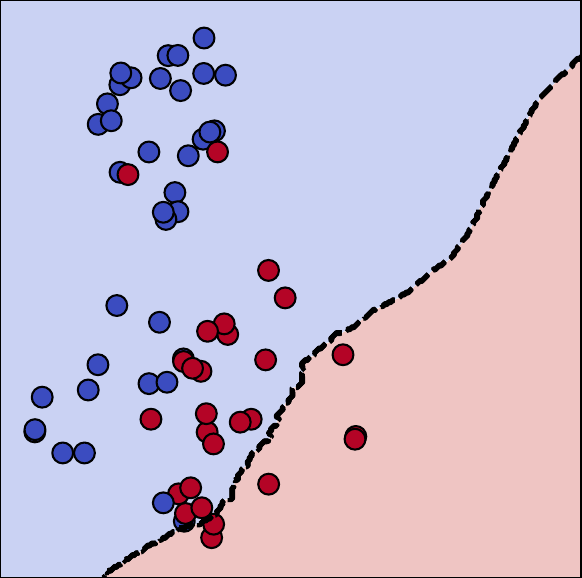}
         \end{subfigure}
         \hfill
         \begin{subfigure}[b]{0.192\textwidth}
             \centering
             After 550 steps\\
             \includegraphics[width=\textwidth]{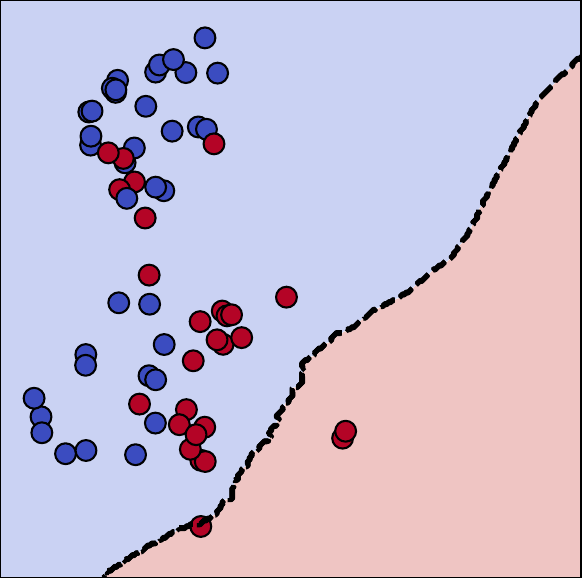}
         \end{subfigure}
         \hfill
         \begin{subfigure}[b]{0.192\textwidth}
             \centering
             After 1100 steps \\
             \includegraphics[width=\textwidth]{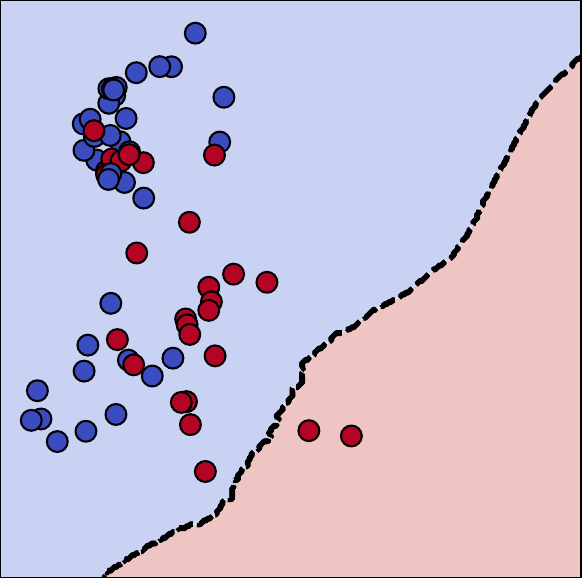}
         \end{subfigure}
         \hfill
         \begin{subfigure}[b]{0.192\textwidth}
             \centering
             After 1650 steps \\
             \includegraphics[width=\textwidth]{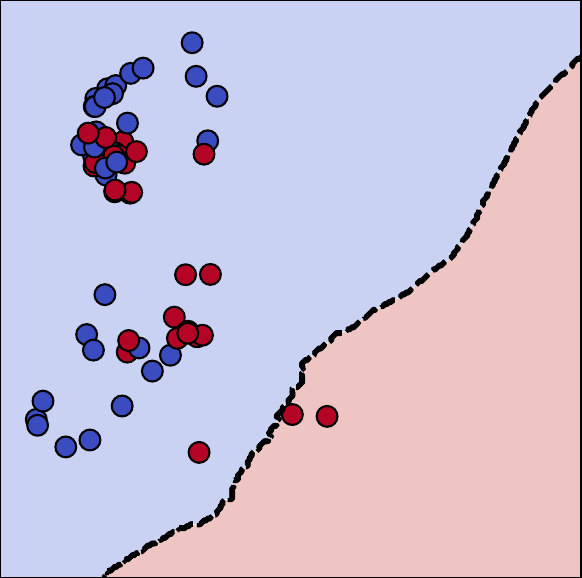}
         \end{subfigure}
         \hfill
         \begin{subfigure}[b]{0.192\textwidth}
             \centering
             After 2200 steps \\
             \includegraphics[width=\textwidth]{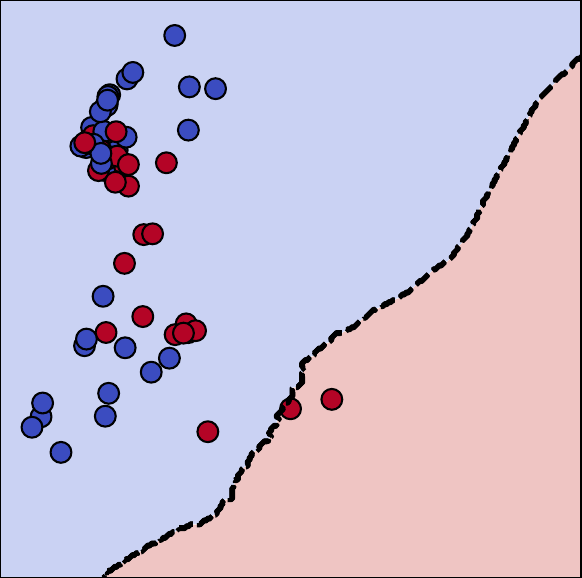}
         \end{subfigure}
         \begin{subfigure}[b]{\textwidth}
             \centering
             \includegraphics[width=\textwidth]{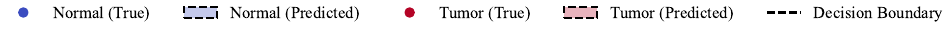}
         \end{subfigure}
     \end{subfigure}
     
        \caption{\textbf{Entropy minimization amplifies prediction bias and ultimately causes model collapse.} We show 2D UMAP embeddings of WILDS-Camelyon features from a ResNet50-GN pretrained on Hospital 2, before and during adaptation. Background colors denote predicted regions in feature space, obtained by fitting a $k$-nearest neighbors classifier to the embeddings and model predictions. As entropy minimization sharpens biased predictions, samples are increasingly reassigned to the dominant class, leading to cluster merging and collapse.}
        \label{fig:tent}
\end{figure}

Table~\ref{tab:pred_bias} illustrates how prediction bias leads to model collapse. Under mild shift, the pre-adaptation predicted distribution is only moderately distorted, and entropy minimization does not collapse the model. Under severe shift, however, the model is already strongly biased before adaptation begins, and entropy minimization amplifies this skew into a nearly single-class solution. Table~\ref{tab:pred_bias} therefore shows that collapse does not emerge from an initially balanced state, but from a sufficiently biased initialization.

Entropy minimization does not create this bias, but amplifies it. By sharpening already dominant predictions and tightening entangled clusters, it pulls ambiguous samples further into overrepresented classes, thereby reinforcing the existing skew. As this feedback loop intensifies, the model collapses to a trivial dominant-class solution. In this sense, prediction bias is not a byproduct of collapse, but its root cause. From Table~\ref{fig:domain_shift_bias}, the ImageNet-C Fog example illustrates this mechanism concretely. Before adaptation, the model already assigns 25.5\% of samples to \emph{Geyser}, despite its true frequency being only 0.2\%, and after entropy minimization, this class absorbs nearly all probability mass. A plausible explanation is that fog introduces appearance statistics resembling those of geysers, displacing many samples toward this region in feature space before adaptation.

\subsection{Why Entropy Minimization Amplifies Prediction Bias}\label{sec:em_amplifies}

In the failure case shown in Fig.~\ref{fig:tent}, more than half of the tumor samples have crossed the decision boundary into the \emph{Normal} cluster. EM treats this merged region as a single cluster and refines it on one side of the boundary, which is precisely how prediction bias amplifies into collapse. This behavior follows from the geometry of entropy minimization. Because EM-based test-time adaptation typically updates only the affine parameters of normalization layers, the classifier $h$ and the decision boundary $\mathcal{H}$ remain fixed throughout adaptation. EM therefore operates entirely within a fixed decision geometry, pushing representations $\phi(x)$ toward more confident predictions. Since the entropy loss is continuous, nearby samples receive similar gradients, so EM refines groups of similarly predicted samples rather than correcting each sample independently.

When the distribution shift has already caused true-class clusters to overlap across $\mathcal{H}$, EM tightens this merged region instead of separating it. Ambiguous samples are pulled further toward the dominant predicted class, $p_{\hat Y}(k)$ becomes increasingly skewed, and each subsequent update starts from an even more biased state. Entropy minimization and prediction bias thus form a positive feedback loop, with model collapse as its endpoint.

\section{Proposed Method: Distribution Shift Bias Reduction (DSBR) }\label{sec:dsbr}

\subsection{Balanced Target Distribution}

To counteract the effect of prediction bias leading to model collapse during EM, we define a \textit{balanced target distribution} that assigns equal probability mass to each predicted cluster while preserving the within-cluster relative density:
\begin{equation}\label{eq:target_dist}
    p_X^*(x)
    \;=\;
    \frac{p_X(x)}{p_{\hat Y}\!\left(\hat y(x)\right)} \cdot \frac{1}{K}
\end{equation}
which is derived from $p_X(x) = p_{X,\hat Y}\!\left(x \mid \hat y(x)\right) \cdot p_{\hat Y}\!\left(\hat y(x)\right)$ by replacing the potentially biased $p_{\hat Y}\!\left(\hat y(x)\right)$ with the uniform distribution $1/K$.
Minimizing the expected loss under $p^*_X$ rather than $p_X$ prevents any single cluster from dominating the gradient signal. 
Because EM's dynamics are gradient-driven, redistributing gradient mass away from dominant clusters provides a corrective force that directly opposes the merging dynamic: samples within an overrepresented cluster, many of which were incorrectly absorbed from other classes, contribute less to each update, while samples in underrepresented clusters, precisely those that resisted incorrect reassignment, contribute more. 
This rebalancing actively encourages the separation of merged clusters rather than their further consolidation. 

At test-time, we cannot sample from $p^*_X$, only from $p_X$. We recover the target loss
$\mathcal{L}^*$ via importance reweighting
\citep{shimodaira2000improving}:
\begin{align}\label{eq:iw_loss}
    \mathcal{L}^* := \mathbb{E}_{X \sim p^*_X}\!\left[l(X)\right]
    = \mathbb{E}_{X \sim p_X}\!\left[\frac{p^*_X(X)}{p_X(X)}\,l(X)\right] = \mathbb{E}_{X \sim p_X}\!\left[\frac{l(X)}{K \cdot p_{\hat Y}\!\left(\hat y(X)\right)}\,\right]
\end{align}
The importance weight is large for samples in underrepresented clusters and small for those in
overrepresented ones.

\subsection{Approximating the Cluster Probabilities at Test-Time}\label{sec:approximating}

The weights in \cref{eq:iw_loss} require knowledge of $p_{\hat Y}(k)$, which is unknown at test-time. Given a batch $\mathcal{B}_t = \{X_1,\ldots,X_B\} \sim p_X(x)$ of size $B$ and sampled at adaptation step $t$, a natural approximation for $p_{\hat Y}(k)$ at adaptation step $t$, which we refer to as $P_t^{(k)}$, is the empirical cluster frequency
\begin{equation}\label{eq:approx_pred_distribution}
    \tilde P_t^{(k)} = \frac{B^{(k)}_t}{B},
    \qquad B^{(k)}_t := \bigl|\{X \in \mathcal{B}_t \mid \hat{y}(X) = k\}\bigr|.
\end{equation}

For datasets with fewer classes, a single-batch estimate is sufficient. For datasets with a high number of classes, such as ImageNet-C ($K{=}1000$), however, batches with $B \ll K$ omit most classes, making per-batch estimates degenerate. Increasing the batch size would address this, but is inefficient. Instead, assuming adaptation steps are small so that $P_t^{(k)} \approx P_{t-1}^{(k)}$, we maintain an exponential moving average (EMA) of cluster frequencies across batches:
\begin{equation}\label{eq:ema}
    \hat{P}_t^{(k)} = \alpha\,\hat{P}_{t-1}^{(k)}
    + (1-\alpha)\,\frac{B^{(k)}_t}{B},
\end{equation}
initialized at $\hat{P}_0^{(k)} = 1/K$ and with decay $\alpha = 0.9$.
The initialization reflects the prior that clusters are balanced before any evidence is observed. Substituting into \cref{eq:iw_loss}, the DSBR loss estimate over batch
$\mathcal{B}_t$ is
\begin{equation}\label{eq:dsbr_loss}
    \hat{\mathcal{L}}^*_t = \frac{1}{B}\sum_{X\in\mathcal{B}_t}
    \frac{l(X)}{K \cdot \hat{P}_t^{(\hat{y}(X))}}.
\end{equation}
We provide the detailed algorithm in Appendix~\ref{sec_app:additional_implementation_details}.

\section{Experiments}\label{sec:experiments}

\textbf{Datasets.} We evaluate on four medical datasets and one natural-vision dataset each with multiple domains and spanning microscopy, X-ray, and MRI, with both 2D and 3D inputs and diverse anatomical sites. \textit{WILDSCamelyon} \citep{koh2021wilds}, derived from Camelyon17 \citep{sun2022camelyon}, is a breast histopathology dataset with 455{,}954 binary classification patches from five hospital domains differing in staining, scanners, and lab procedures. \textit{Histopantum} \citep{zamanitajeddin2024benchmarking} is a binary tumor histopathology dataset on TCGA-derived patches \citep{weinstein2013cancer}, with four domains corresponding to colorectal, ovarian, stomach, and uterine cancers, yielding biological rather than instrumental shift. \textit{MammoBench} \citep{bhole2025mammobench} unifies six mammography datasets (domains) \citep{moreira2012inbreast, kopans2020ddsm, alsolami2021king, khaled2021categorized, Oza2023, cui2021chinese} into a Normal-vs-Abnormal dataset; excluding CMMD \citep{cui2021chinese}, which lacks Normal cases, leaves 14{,}529 images across the remaining domains. \textit{GliomaMRI} follows \citet{scholz2024imbalance} and combines UCSF-PDGM \citep{calabrese2022university}, EGD \citep{van2021erasmus}, and TCGA \citep{bakas2017advancing} for three-class glioma subtype classification under the 2021 WHO CNS taxonomy \citep{louis2021who}; with 1{,}146 volumes, it is the smallest dataset and the only 3D one. \\
We use \textit{ImageNet-C} \citep{hendrycks2019benchmarking}, which comprises 75 corruption domains across 15 corruption types and 5 severity levels on ImageNet \citep{deng2009imagenet}, to study the high-$K$ regime. Moreover, we use this dataset with multiple challenging scenarios. We provide additional dataset details in Appendix~\ref{sec_app:datasets}.

\paragraph{Implementation details.} We use two backbones for all 2D datasets: ResNet50-GN~\citep{he2016deep, wu2018group} and ViT-B/16~\citep{dosovitskiy2020image}. Following seminal works~\cite{niu2023towards,lee2024entropy}, we use ResNet's Group Normalization model to avoid known issues of entropy minimization, such as limited batch sizes, and use the same model for all our ablations. For GliomaMRI, we use volumetric architectures from the MONAI framework~\citep{cardoso2022monai}: a 3D ResNet10 and a 3D ViT-T/16. \begin{wrapfigure}{r}{0.5\linewidth} %
  \centering
  \includegraphics[width=0.89\linewidth]{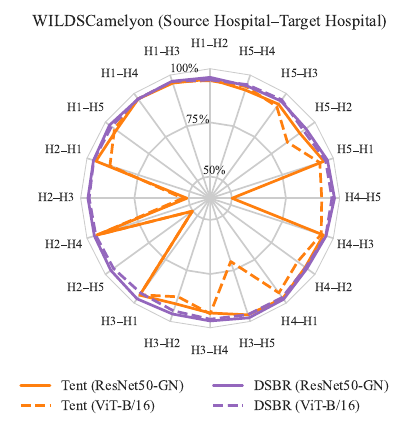}
    \caption{\textbf{DSBR stabilizes Entropy Minimization} where `H' denotes the Hospital domain id. Balanced accuracy results on all \textit{WILDSCamelyon} configurations for Tent (EM) and DSBR.}
    \label{fig:radar}
  \vspace{-15pt}
\end{wrapfigure}

 For source-target, i.e., training and adaptation data pairing on each medical dataset, every pair of distinct domains forms one configuration, except for MammoBench and GliomaMRI, where the small training sets motivate leave-one-domain-out pre-training instead.  We use a batch size of 32 for pre-training and adaptation on the medical datasets, and 64 on ImageNet-C, matching prior work~\cite{lee2024entropy,niu2023towards,wang2021tent}. In line with \citep{press2024entropy}, for all baselines and our method, we hold out a 20\% split that is excluded from adaptation and used solely for evaluation in each medical domain since online accuracy can mask late-stage collapse and the final adapted state is what matters for clinical deployment. For ImageNet-C we follow prior work and adapt on the full corruption domain without a holdout split. We compare DSBR against four state-of-the-art EM-based baselines, all re-implemented for the medical datasets. Throughout, all reported accuracies are balanced accuracies; ROC-AUC results are reported in Appendix~\ref{app:roc_auc}. DSBR uses an EMA decay of $\alpha = 0.9$ unless specified otherwise. We provide additional details in the appendix. 

\begin{table}[]
    \centering
    \resizebox{1.0\linewidth}{!}{
    \begin{threeparttable}
    \caption{\textbf{Balanced accuracy (\%) on the four medical benchmarks.} The results are averaged over all source-target configurations of each dataset. We re-implement the baselines on the medical datasets and report the average improvement over the corresponding non-adapted baseline (ResNet, ViT) as $\Delta$~Average.
    \label{tab:results_medical}
    \textbf{Bold} marks the best result per column, and \underline{underlined} the second-best. DSBR is either the top performer or performs competitively}
        \begin{tabular}{lccccc}
        \toprule
        Model \& Method & \textbf{Histopantum} & \textbf{MammoBench} & \textbf{WILDSCamelyon} & \textbf{GliomaMRI} & \textbf{$\Delta$} \textbf{Average} \\
        \midrule
        ResNet & 86.6$_{\pm0.0}$ & 59.4$_{\pm0.0}$ & 87.4$_{\pm0.0}$ & 42.9$_{\pm0.0}$ & \neu0.0$_{\pm0.0}$ \\
        ~~$\bullet~$Tent~\cite{wang2021tent} & \underline{88.5$_{\pm0.1}$} & 58.8$_{\pm0.1}$ & 87.9$_{\pm0.0}$ & \underline{44.3$_{\pm1.3}$} & \pos0.8$_{\pm0.4}$ \\
        ~~$\bullet~$SAR~\cite{lee2023towards} & 86.6$_{\pm0.0}$ & 59.4$_{\pm0.0}$ & 87.4$_{\pm0.0}$ & 44.2$_{\pm1.4}$ & \pos0.3$_{\pm0.4}$ \\
        ~~$\bullet~$ROID~\cite{marsden2024universal} & 87.2$_{\pm1.7}$ & 56.6$_{\pm0.5}$ & \underline{94.8$_{\pm0.6}$} & \textbf{44.4$_{\pm1.3}$}$^*$ & \underline{\pos1.7$_{\pm1.0}$} \\
        ~~$\bullet~$DeYO~\cite{lee2024entropy} & 76.1$_{\pm2.8}$ & \underline{60.9$_{\pm0.2}$} & 92.7$_{\pm0.5}$ & n/a$^\dagger$ & \nega0.9$_{\pm0.9}$ \\
        \rowcolor{DSBR_purple!20}~~$\bullet~$\textbf{\textit{DSBR}} (ours) & \textbf{89.9$_{\pm0.4}$} & \textbf{62.7$_{\pm0.5}$} & \textbf{96.5$_{\pm0.1}$} & 44.0$_{\pm1.1}$ & \textbf{\pos4.2$_{\pm0.5}$} \\
        \midrule
        ViT & 87.1$_{\pm0.0}$ & 54.4$_{\pm0.0}$ & 87.0$_{\pm0.0}$ & 37.2$_{\pm0.0}$ & \neu0.0$_{\pm0.0}$ \\
        ~~$\bullet~$Tent~\cite{wang2021tent} & 87.4$_{\pm0.0}$ & 54.4$_{\pm0.0}$ & 87.4$_{\pm0.1}$ & \underline{37.4$_{\pm0.2}$} & \pos0.2$_{\pm0.1}$ \\
        ~~$\bullet~$SAR~\cite{lee2023towards} & 87.1$_{\pm0.0}$ & 54.4$_{\pm0.0}$ & 87.0$_{\pm0.0}$ & 37.2$_{\pm0.0}$ & \neu0.0$_{\pm0.0}$ \\
        ~~$\bullet~$ROID~\cite{marsden2024universal} & \underline{88.0$_{\pm0.3}$} & \underline{54.5$_{\pm0.8}$} & \underline{94.3$_{\pm0.2}$} & \textbf{38.2$_{\pm5.0}$}$^*$ & \underline{\pos2.3$_{\pm1.6}$} \\
        ~~$\bullet~$DeYO~\cite{lee2024entropy} & 87.9$_{\pm0.2}$ & \textbf{55.2$_{\pm0.1}$} & 93.5$_{\pm0.3}$ & n/a$^\dagger$ & \pos2.0$_{\pm0.2}$ \\
        \rowcolor{DSBR_purple!20}~~$\bullet~$\textbf{\textit{DSBR}} (ours) & \textbf{88.8$_{\pm0.4}$} & \textbf{55.2$_{\pm0.5}$} & \textbf{95.6$_{\pm0.3}$} & 36.5$_{\pm0.9}$ & \textbf{\pos2.6$_{\pm0.5}$} \\
        \bottomrule
        \end{tabular}

        \end{threeparttable}
    }
    
\end{table}

\subsection{Eliminating Model Collapse for Medical Imaging}\label{sec:results_medical}
In \cref{tab:results_medical}, we provide the balanced accuracy on the medical datasets averaged over all source-target configurations. DSBR achieves the highest average improvement on both backbones, $+4.2\%$ for ResNet and $+2.6\%$ for ViT, considerably more than the next-best baseline (ROID, $+1.7\%$ and $+2.3\%$). On individual datasets, DSBR is the top performer on Histopantum, MammoBench, and WILDSCamelyon for both backbones, and within $0.4\%$ of the best on GliomaMRI. The only non-leading entries are on GliomaMRI, where ROID is best with its consistency component disabled, and DeYO is inapplicable, since both depend on 2D augmentations that do not transfer to 3D MRI volumes. \bluetext{We attribute DSBR’s limited gains on the GliomaMRI dataset to the low performance of the source model, which results from the small training set and also limits the gains of all baseline methods. Nevertheless, despite these small gains, DSBR remains within the error margins of the best-performing methods for both backbones.}

From ~\cref{tab:results_medical}, the dataset-level averages do not reveal the instability of EM. \purpletext{Therefore, in figure }\Cref{fig:radar} \purpletext{we report} per-configuration accuracy on WILDSCamelyon for Tent and DSBR, \purpletext{which reveals Tent's tendency towards model collapse.} In our single-source pairing, Tent collapses to chance-level accuracy ($\approx 50\%$) on a substantial fraction of hospital pairs for both backbones, while DSBR shows no such collapses: every configuration attains near-optimal accuracy (visible as the outer ring in \cref{fig:radar}). Targeting prediction bias does not just improve average accuracy; it eliminates the configurations in which Tent fails outright.

\begin{table}[t!]
    \centering
    \small
    \caption{\textbf{Accuracy (\%) on ImageNet-C on default and wild settings \cite{niu2023towards} at severity level 5.}
    Results are averaged over 15 corruption types, except \emph{Mixed Corruptions} where one run comprises a mixture of all 15 corruptions. We report the average improvement over the corresponding non-adapted baseline as $\Delta$~Average. DSBR is best \textbf{(bold)} or competitive across settings and is the top performer on average. }
    \begin{tabular}{lccccc}
    \toprule
     Model \& Method & \textbf{Default} & \textbf{Label Shifts} & \textbf{Mixed Corruptions} & \textbf{Batch Size 1} & $\Delta$ \textbf{Average} \\
     \midrule
    ResNet50-GN & 31.4$_{\pm0.0}$ & 31.4$_{\pm0.0}$ & 31.4$_{\pm0.0}$ & 31.4$_{\pm0.0}$ & \neu0.0$_{\pm0.0}$ \\
    ~~$\bullet~$Tent~\cite{wang2021tent}   & 32.2$_{\pm0.3}$ & 30.3$_{\pm0.7}$ & 32.1$_{\pm1.7}$ & 30.8$_{\pm0.3}$ & \neu0.0$_{+0.7}$ \\
    ~~$\bullet~$SAR~\cite{niu2023towards}  & 36.7$_{\pm3.0}$ & 37.8$_{\pm0.5}$ & 38.5$_{\pm0.1}$ & 35.8$_{\pm0.3}$ & \pos5.8$_{\pm1.0}$ \\
    ~~$\bullet~$ROID~\cite{marsden2024universal} & 46.8$_{\pm0.1}$ & 2.7$_{\pm0.1}$ & 37.6$_{\pm0.1}$ & 0.1$_{\pm0.0}$ & \nega9.6$_{\pm0.1}$ \\
    ~~$\bullet~$DeYO~\cite{lee2024entropy} & 44.1$_{\pm4.9}$ & \underline{44.1$_{\pm5.6}$} & 39.4$_{\pm1.1}$ & \textbf{43.9$_{\pm4.3}$} & \underline{\pos11.4$_{\pm4.0}$} \\
    \rowcolor{DSBR_purple!20}~~$\bullet~$\textbf{\textit{DSBR}}($\alpha=0.99$) & \textbf{49.7$_{\pm0.1}$} & \textbf{46.8$_{\pm0.2}$} & \textbf{43.0$_{\pm0.2}$} & \underline{41.6$_{\pm0.1}$} & \textbf{\pos13.9$_{\pm0.1}$} \\
    \rowcolor{DSBR_purple!20}~~$\bullet~$\textbf{\textit{DSBR}}($\alpha=0.9$)  & \underline{48.8$_{\pm0.1}$} & 38.2$_{\pm0.3}$ & \underline{41.4$_{\pm0.2}$} & 38.3$_{\pm0.1}$ & \pos10.2$_{\pm0.2}$ \\
    \midrule
    ViT-B/16 & 39.8$_{\pm0.0}$ & 39.8$_{\pm0.0}$ & 39.8$_{\pm0.0}$ & 39.8$_{\pm0.0}$ & \neu0.0$_{\pm0.0}$ \\
    ~~$\bullet~$Tent~\cite{wang2021tent}   & 51.9$_{\pm0.5}$ & 53.0$_{\pm0.3}$ & 33.1$_{\pm4.5}$ & 51.7$_{\pm0.3}$ & \pos7.6$_{\pm1.4}$ \\
    ~~$\bullet~$SAR~\cite{niu2023towards}  & 54.2$_{\pm0.3}$ & 56.2$_{\pm0.1}$ & 53.7$_{\pm0.2}$ & 56.2$_{\pm0.3}$ & \pos15.3$_{\pm0.2}$ \\
    ~~$\bullet~$ROID~\cite{marsden2024universal} & \underline{64.6$_{\pm0.1}$} & \textbf{86.4$_{\pm4.9}$} & 57.2$_{\pm0.1}$ & 0.1$_{\pm0.0}$ & \pos12.2$_{\pm1.3}$ \\
    ~~$\bullet~$DeYO~\cite{lee2024entropy} & 61.8$_{\pm2.0}$ & \underline{62.4$_{\pm1.1}$} & 53.6$_{\pm0.8}$ & \textbf{61.8$_{\pm0.9}$} & \underline{\pos20.1$_{\pm1.2}$} \\
    \rowcolor{DSBR_purple!20}~~$\bullet~$\textbf{\textit{DSBR}}($\alpha=0.99$) & \textbf{64.9$_{\pm0.3}$} & 56.2$_{\pm0.2}$ & \textbf{60.9$_{\pm0.7}$} & \underline{58.2$_{\pm0.1}$} & \textbf{\pos20.2$_{\pm0.3}$} \\
    \rowcolor{DSBR_purple!20}~~$\bullet~$\textbf{\textit{DSBR}}($\alpha=0.9$)  & 64.4$_{\pm0.2}$ & 42.2$_{\pm0.2}$ & \underline{59.5$_{\pm1.0}$} & 55.9$_{\pm0.4}$ & \pos15.7$_{\pm0.5}$ \\
    \bottomrule
    \end{tabular}
    \label{tab:imagenetc_results}
\end{table}

\begin{figure}[ht]
    \centering
    \begin{subfigure}[b]{0.49\linewidth}
    \centering
    \includegraphics[width=\linewidth]{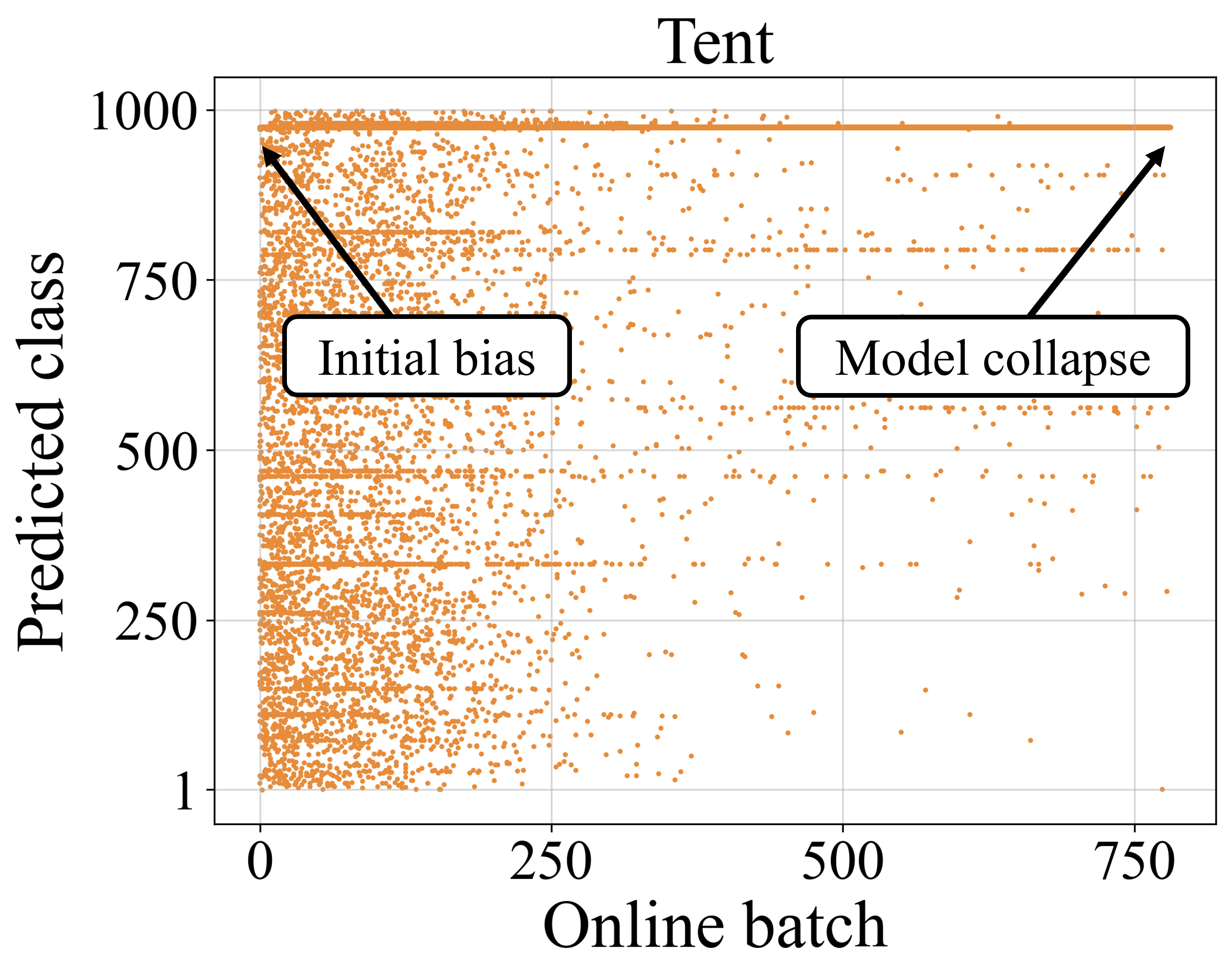}
    \end{subfigure}
    \hfill
    \begin{subfigure}[b]{0.49\linewidth}
    \centering
    \includegraphics[width=\linewidth]{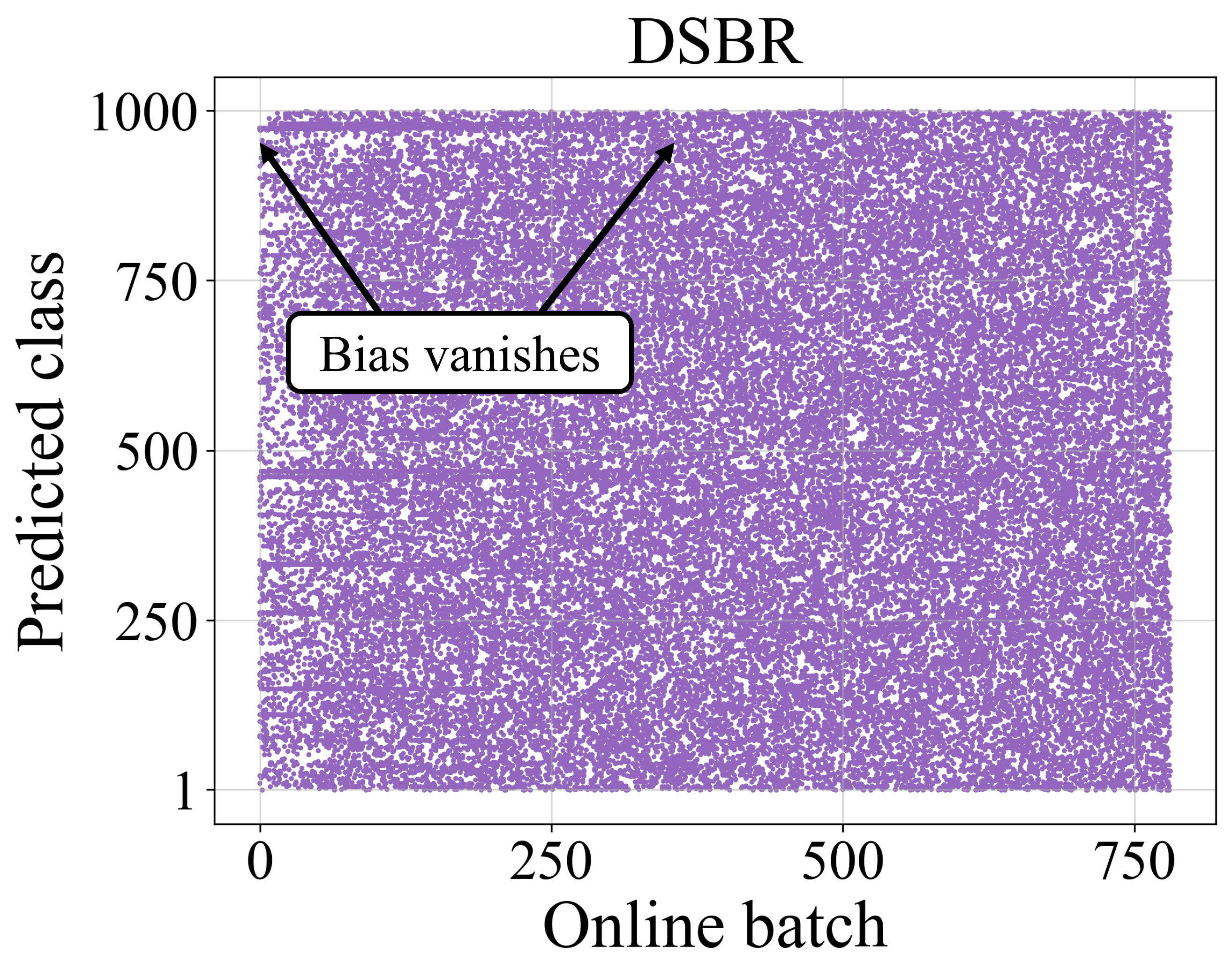}
    \end{subfigure}
    \caption{\textbf{Model collapse in Tent (EM) and its mitigation by DSBR.} Initial bias is present for both models.
    \textbf{Left:} Tent amplifies bias until the model converges to a single class for almost every input (\emph{Model collapse}) by the end of the episode.
    \textbf{Right:} DSBR progressively dampens bias until it vanishes by approximately batch 400.}
    \label{fig:dsbr_vs_tent_predictions}
\end{figure}

\subsection{Performance on natural imaging by mitigating bias}\label{sec:results_imagenet}
\Cref{tab:imagenetc_results} reports results on ImageNet-C in four settings: standard adaptation, online imbalanced label shifts with imbalance ratio $\infty$, mixed corruptions across all 15 types, and batch size 1, all at severity 5 \cite{niu2023towards}. We report DSBR with two EMA decays: the default $\alpha = 0.9$ and a higher $\alpha = 0.99$, motivated by \cref{sec:approximating}: when $K \gg B$ ($K=1000$, $B=64$), the per-batch frequency is sharply peaked on the few classes that appear in the batch, and a higher decay softens it across batches before reweighting. On the standard and mixed-corruption settings, DSBR achieves the highest accuracy on both backbones. 

For label shifts and batch size 1, DSBR with $\alpha = 0.99$ performs better than DSBR with $\alpha = 0.9$ since each batch is dominated by a single class. Moreover, this DSBR $\alpha = 0.99$ performs best on label shifts for ResNet50-GN ($+2.7\%$ over the next-best baseline) and second-best at batch size 1 on both backbones, demonstrating that the EMA preserves cluster-frequency information across batches even when a single batch is uninformative.

\begin{wrapfigure}{r}{0.40\linewidth}   %
    \vspace{-4pt}                          %
    \centering
    \includegraphics[width=\linewidth]{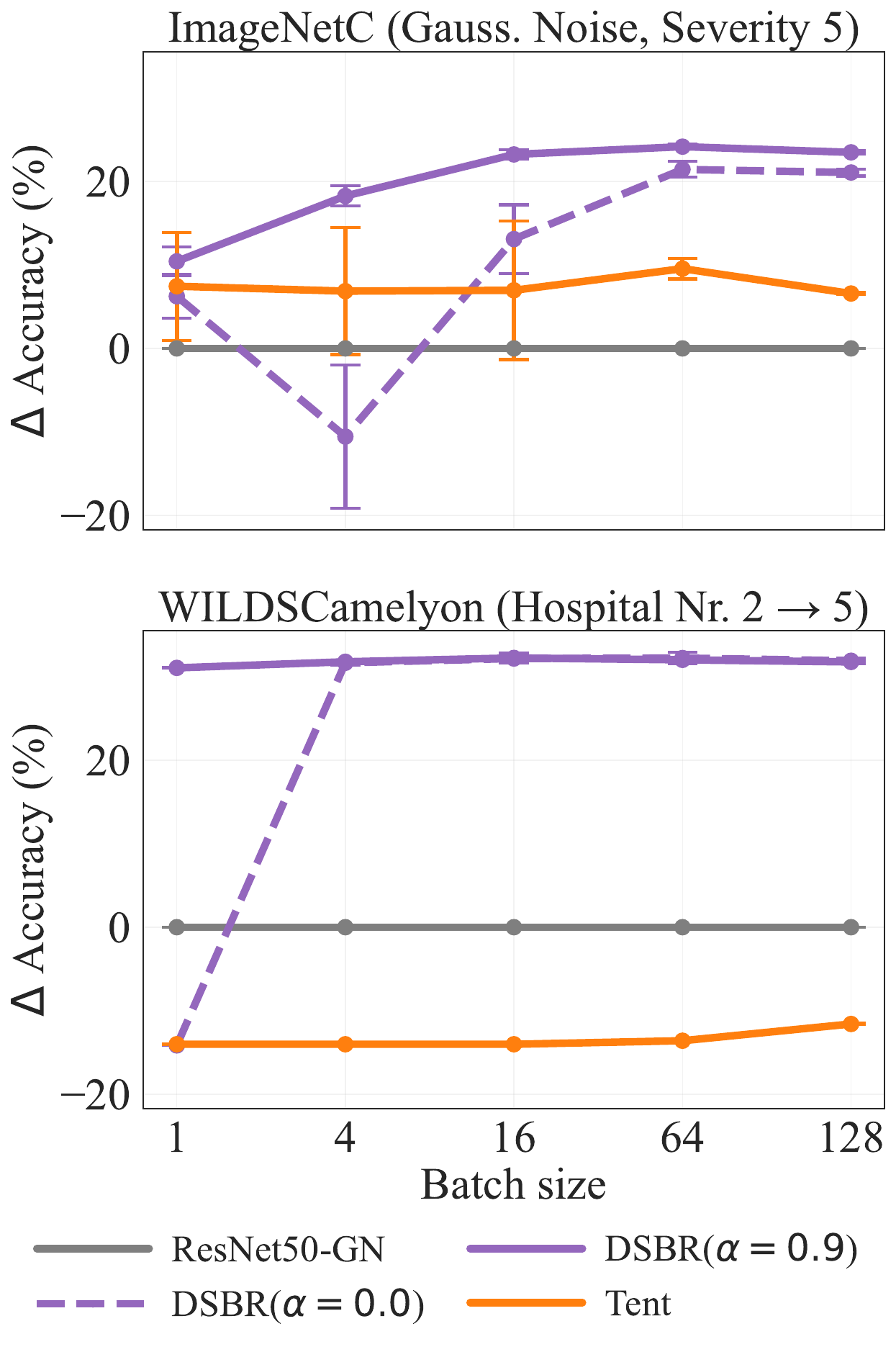}
    \caption{\textbf{EMA stabilizes DSBR for limited batch sizes.}
Balanced accuracy across batch sizes on ImageNet-C and WILDSCamelyon
(ResNet50-GN), evaluated on a held-out 20\% split. DSBR remains
effective for limited batch sizes. }
\label{fig:ablation_bs}
    \vspace{-32pt}
\end{wrapfigure}

\Cref{fig:dsbr_vs_tent_predictions} visualizes the mechanism behind these gains on a single ImageNet-C Fog (severity 5) episode. Both Tent and DSBR start from the same source model and inherit the same noticeable initial prediction bias toward a small set of dominant classes induced by the corruption. Under Tent, this bias is amplified rather than corrected: within a few hundred batches, predictions concentrate on a single class, and by the end of the episode, the model has collapsed. Under DSBR, the same initial bias is progressively dampened by the importance reweighting; by the second half of the episode, the predicted distribution has spread across the full label space and remains spread thereafter.

\subsection{EMA stabilizes DSBR for limited batch sizes}\label{sec:acc_bias_estimation}

We demonstrate the benefits of using EMA for DSBR in \Cref{fig:ablation_bs}. DSBR's effectiveness depends on the quality of the running estimate $\hat{P}_t^{(k)}$ of the predicted class distribution, which in turn depends on the batch size $B$, the number of classes $K$, and the EMA decay $\alpha$. We ablate these by reporting DSBR with $\alpha = 0.9$ and $\alpha = 0$ across batch sizes from 1 to 128, on representative configurations from each end of the $K$ spectrum: ImageNet-C Gaussian Noise at severity 5 ($K=1000$) and WILDSCamelyon Hospital Nr. 2 $\to$ 5 ($K=2$). At batch size 1 and $\alpha = 0$, without EMA, DSBR degrades to standard Tent for both datasets, since the loss reduces to Tent's up to a constant factor of $1/K$. Enabling the EMA recovers near-optimal performance immediately on WILDSCamelyon. For low-$K$ benchmarks, the per-batch estimator alone matches the EMA estimator at moderate batch sizes ($B \geq 4$); for high-$K$ benchmarks, the EMA remains necessary throughout. DSBR's reweighting works straightforwardly when $B$ is on the order of $K$ or larger, and the EMA is what extends it to the regime $B \ll K$.

\section{Conclusion}\label{sec:conclusion}

 We identify the root cause of entropy‑minimization (EM) failure under distribution shift: the merging of feature‑space clusters. 
The merging induces prediction bias, and EM, which refines whatever clusters are present in feature space, treats the merged region as a single dominant cluster and further compresses it. This reinforces the bias and ultimately results in a single-class collapse.
To address this failure, we propose Distribution Shift Bias Reduction (DSBR), which balances predicted-class contributions during adaptation. In contrast to other EM state-of-the-art methods, DSBR does not require sample filtering or domain-specific augmentations and uses only a single hyperparameter (the EMA decay). In addition, DSBR operates solely at test time. 

We extensively evaluate DSBR across medical‑imaging and natural-vision datasets. Across all settings, DSBR consistently stabilizes test‑time adaptation, prevents model collapse, and performs competitively or outperforms current state‑of‑the‑art methods. DSBR depends on an accurate estimate of the predicted‑class distribution. This estimate may deteriorate in a few specific cases, such as imbalanced label shifts or extremely small batch sizes relative to the number of classes. 
However, our experiments demonstrate that correcting a single, mechanistically grounded quantity, the prediction bias that arises from cluster merging is sufficient to stabilize entropy minimization on both medical‑imaging and natural‑image benchmarks. This provides strong empirical evidence that cluster merging (as reflected by prediction bias) is the primary objective to address when stabilizing EM. More broadly, our findings highlight the importance of a principled mechanistic understanding for developing robust and reliable test-time adaptation methods under distribution shift.

\section*{Acknowledgments}
This work is supported by DAAD programme Konrad Zuse Schools of Excellence in Artificial Intelligence, Munich Center for Machine Learning (MCML), both sponsored by the Federal Ministry of Research, Technology, Bavaria California Technology Center (BaCaTeC, Bayerisch-Kalifornischen Hochschulzentrum) under BaCaTeC-Förderprojekt Nr. 20 [2026-1], Deutsche Forschungsgemeinschaft (DFG, German Research Foundation) – 515279324 / SPP 2177, and the German Federal Ministry of Research, Technology and Space (DECIPHER-M, 01KD2420G). DML and JAS received funding from HELMHOLTZ IMAGING, a platform of the Helmholtz Information and Data Science Incubator. 
\newpage

\nocite{shah2020predictive}
\nocite{su2023beware}
\nocite{bar2024protected}

\nocite{chen2022pathologies}
\nocite{lin2024optimizing}
\nocite{guo2024bias}

\nocite{agarwal2026the}
\nocite{kazdan2024collapse}
\nocite{lin2024optimizing}

\nocite{dohmatob2025strong}
\nocite{schaeffer2025position}
\nocite{gambetta2025learningsurprisesurplexitymitigating}

\nocite{shi2025closer}
\nocite{dang2023neural}
\nocite{zhong2023understanding}

\nocite{chung2024mitigatingbiasmodelcontinual}

\nocite{mounsaveng2024bag}

\nocite{seto2023realmrobustentropyadaptive}

\nocite{zhao2023pitfalls}
\nocite{darestani2022test}

\nocite{maharana2026continual}
\nocite{kim2024testtime}
\nocite{goyal2022test}

\nocite{liang2025comprehensive}
\nocite{wang2023search}

\nocite{su2024towards}
\nocite{han2025ranked}
\nocite{ambekar2026mean}

\bibliography{neurips_2026, references}
\bibliographystyle{abbrvnat}

\newpage
\appendix

\startcontents[appendix]
\printcontents[appendix]{}{1}{%
    \section*{Appendix}
}
\newpage

\section{Additional Implementation Details}\label{sec_app:additional_implementation_details}

Our method's code is based on Domainbed~\cite{gulrajani2021in} and is efficient. We pre-train all source models using a unified DomainBed-based pipeline \citep{gulrajani2021in}, modified to allow for early stopping when validation losses stop decreasing and gradient accumulation for 3D volumes where no full batch fits into VRAM.
Each domain is split 80/20 into IN and OUT subsets.
Pre-training uses source-IN for training and source-OUT for validation, while adaptation uses target-IN as the stream of inputs the model adapts on and target-OUT as the held-out evaluation set on which we report performance.
Evaluating on a held-out target split rather than online during adaptation is a deliberate choice, because online accuracy can mask late-stage collapse: a model can produce reasonable predictions for the first several thousand samples and then collapse to a trivial constant predictor in the final thousand, leaving the running average of online accuracy looking acceptable while the model is in fact unusable.
In clinical settings, where the input stream has no natural endpoint, we need to be confident that the model does not degrade at any point during deployment, which makes the \emph{final} adapted state more informative than the online accuracy.
We use batch size 32 on the medical datasets (constrained by VRAM for the volumetric inputs) and 64 on ImageNet-C, matching prior work.
For ImageNet-C we follow prior-work convention instead \citep{wang2021tent, niu2022efficient, niu2023towards, lee2024entropy}: we use publicly available ImageNet-pretrained checkpoints as source models and adapt on the full target corruption domain without a held-out split.
All reported numbers are averaged over 5 seeds for the medical datasets and 3 seeds for ImageNet-C; 2 standard deviations across seeds are calculated using \emph{pandas} library std function and reported alongside each mean.

For the re-implementation of related work's baselines~\cite{wang2021tent,niu2023towards,marsden2024universal,lee2024entropy} we stick closely to the official repositories and use the default hyperparameters and optimizers of each method.

\begin{algorithm}[ht]
\small
\caption{\textbf{D}istribution \textbf{S}hift \textbf{B}ias \textbf{R}eduction (DSBR)} 
{\textbf{Input:}} Test samples $\mathcal{X}=\{x_i\}_{i=1}^{N^{\mathrm{test}}}$, 
model $f_\theta(x)$ with trainable parameters $\tilde{\theta}\subset\theta$, 
batch size $B$, optimizer $\mathcal{O}_\eta(\cdot)$ with learning rate $\eta>0$, 
EMA decay $\alpha$\\
\textbf{Output:} Predictions $\{\hat{y}_i\}_{i=1}^{N^{\mathrm{test}}}$, adapted parameters $\tilde{\theta}$.
\label{alg:DSBR}
\begin{algorithmic}[1]
\STATE Initialize model parameters $\tilde{\theta}$ with a chosen scheme.
\STATE Initialize $\hat{P}^{(k)} \gets 1/K$ for all $k \in \{1,\dots,K\}$
\FOR{$b = 1$ to $N^\text{test}$ step $B$}
    \STATE $\mathcal{B} \gets \{ b, \dots, \min(b+B-1, N) \}$
    \STATE $\hat{y}_i \gets \arg\max_j\, f_\theta(x_i)_j$ \textbf{for} every $i\in\mathcal{B}$
    \STATE $B^{(k)} \gets |\{ i \in \mathcal{B} \mid \hat{y}_i = k\}|$ \textbf{for} every $k \in \{1,\dots,K\}$ 
    \STATE $\hat{P}^{(k)} \gets \alpha\,\hat{P}^{(k)} + (1-\alpha)\,{B^{(k)}}/{|\mathcal{B}|}$ \textbf{for} every $k \in \{1,\dots,K\}$ 
    \hfill $\triangleright$ \cref{eq:ema}
    \STATE $l_i \gets H(f_\theta(x_i))$ \textbf{for} every $i \in \mathcal{B}$  
    \STATE $\mathcal{L} \gets \dfrac{1}{B}\displaystyle\sum_{i\in\mathcal{B}} \dfrac{l_i}{K \cdot \hat{P}^{(\hat{y}_i)}}$ \hfill $\triangleright$ \cref{eq:dsbr_loss}
    \STATE Update $\tilde{\theta} \gets \mathcal{O}_\eta\!\left(\tilde{\theta},\, \nabla_{\tilde{\theta}}\mathcal{L}\right)$
\ENDFOR
\end{algorithmic}
\end{algorithm}

\textbf{Full Algorithm and Optimizer. }\cref{alg:DSBR} provides the pseudo-code for our method. As optimizer $\mathcal{O}_\eta$ we use Adam \cite{kingma2014adam} with default hyperparameters:
$\beta_1 = 0.9$, $\beta_2 = 0.999$, $\epsilon = 10^{-8}$, and $\lambda = 0$ (no weight decay).
For the learning rate $\eta$ we use the same per batch size defaults as \citet{wang2021tent}.

\paragraph{Comparison with results from other works.} For the ImageNet-C datasets, our re-implementation obtains higher results than the results reported in baseline methods. We verified that this is due to the usage of pip packages instead of conda. We encourage the community to adopt pip to implement Tent for baselines.

\section{Additional Datasets Details}
\label{sec_app:datasets}
 
We evaluate on four medical datasets and one natural vision dataset.
We do not cover the full breadth of medical imaging in this evaluation, but the four datasets below span a substantial cross-section of it: three modalities (microscopy, X-ray, and MRI), two- and three-dimensional inputs, and several anatomical sites (breast, brain, and a range of organs covered by the histopathology datasets). \bluetext{Notably, these datasets exhibit real-world distribution shifts and the class imbalance typical of the medical domain.
We quantify the latter using the imbalance ratio $\mathrm{IR} = N_{\text{majority}}/N_{\text{minority}}$. We report, for each dataset, the maximum IR across its domains, reflecting the most skewed setting a model may encounter. GliomaMRI is the most imbalanced ($\mathrm{IR} = 28.31$), followed by MammoBench ($\mathrm{IR} = 5.15$) and Histopantum ($\mathrm{IR} = 1.65$), whereas WILDS-Camelyon is the only balanced dataset ($\mathrm{IR} = 1.00$).
}
 
\textbf{WILDSCamelyon} \citep{koh2021wilds} is a histopathology dataset for breast cancer metastasis detection, derived from Camelyon17 \citep{sun2022camelyon} by extracting patches from whole-slide images for binary cancerous-vs-noncancerous classification; the five domains correspond to five hospitals, each with its own staining protocol, scanner, and lab procedure, totaling 455{,}954 patches. {License: Creative Commons Zero (CC0)}
 
\textbf{Histopantum} \citep{zamanitajeddin2024benchmarking} is a histopathology dataset for binary tumor classification on patches derived from TCGA \citep{weinstein2013cancer}, where the four domains correspond to four cancer types: colorectal, ovarian, stomach, and uterus; adaptation across these domains is biological rather than instrumental. The dataset comprises 281{,}142 patches sourced from whole-slide images annotated by experienced pathologists {License: Creative Commons Attribution-NonCommercial-ShareAlike 4.0 International (CC BY-NC-SA 4.0)}
 
\textbf{MammoBench} \citep{bhole2025mammobench} aggregates six mammography datasets \citep{moreira2012inbreast, kopans2020ddsm, alsolami2021king, khaled2021categorized, Oza2023, cui2021chinese} into a unified Normal-vs-Abnormal dataset; we exclude CMMD \citep{cui2021chinese} because it does not contain Normal cases, leaving 14{,}529 images across the remaining datasets, each treated as a domain. The dataset underwent rigorous preprocessing including breast segmentation, pectoral muscle removal, and intelligent cropping to ensure consistency across images from diverse sources while preserving clinically relevant features. {License: Creative Commons Attribution-NonCommercial-ShareAlike 4.0 International (CC BY-NC-SA 4.0)}
 
\textbf{GliomaMRI} follows \citet{scholz2024imbalance} and combines three 3D brain MRI datasets (UCSF-PDGM \citep{calabrese2022university}, EGD \citep{van2021erasmus}, and TCGA \citep{bakas2017advancing}) for three-class glioma subtype classification under the 2021 WHO classification of CNS tumors \citep{louis2021who}, with the three classes being IDH-wildtype glioblastoma, IDH-mutant 1p/19q-intact astrocytoma, and IDH-mutant 1p/19q-codeleted oligodendroglioma; with 1{,}146 volumes it is the smallest dataset and the only one with volumetric inputs. The UCSF-PDGM dataset includes 501 subjects with histopathologically-proven diffuse gliomas imaged with a standardized 3 Tesla preoperative brain tumor MRI protocol featuring predominantly 3D imaging and advanced diffusion and perfusion imaging techniques. {License: TCIA Data Usage Policy (open access with attribution); TCGA: NIH Genomic Data Sharing Policy (open and controlled access tiers); EGD: Open access}
 
\textbf{ImageNet-C} \citep{hendrycks2019benchmarking} provides 75 corruption domains, each consisting of one of 15 corruption types at one of 5 severity levels, applied to the ImageNet \citep{deng2009imagenet} validation set, with 50{,}000 images and 1{,}000 classes per domain; we use it to compare DSBR against prior TTA work calibrated on this dataset and to study the high-$K$ regime. The corruption types span noise, blur, weather, digital, and additional degradation categories to comprehensively evaluate classifier robustness. {License: Creative Commons Attribution 4.0 International (CC BY 4.0) (for corruptions); ImageNet base dataset: Non-commercial research and educational use}

\section{Pre-training Results}\label{sec_app:pre-training}
We perform pre-training on four multi-domain medical imaging datasets containing 5, 4, 3, and 5 domains for \textbf{WILDSCamelyon}, \textbf{Histopantum}, \textbf{GliomaMRI}, and \textbf{MammoBench}, respectively. The strategy for selecting training domains differs by dataset. For \textbf{WILDSCamelyon} and \textbf{Histopantum}, we use each individual domain as a separate training set, since these datasets are large enough to produce reasonable models from a single source. This also matches the realistic scenario in which a practitioner has access to data from only one institution. For \textbf{GliomaMRI} and \textbf{MammoBench}, the per-domain data is too small for stand-alone training, so we instead train on every leave-one-out combination of domains. With two backbone architectures evaluated per dataset, this yields 10, 8, 6, and 10 pretrained models per dataset, for a total of \textbf{34 pretrained models}.

Per-dataset results are reported in Tables~\ref{tab:pre_wildscamelyon}, \ref{tab:pre_histopantum}, \ref{tab:pre_glioma}, and \ref{tab:pre_mammobench}. The histopathology datasets, \textbf{WILDSCamelyon} and \textbf{Histopantum}, yield strong pretrained models: their Avg.~Source OUT scores approach 100\%, indicating that the models reliably fit the source distribution. In contrast, \textbf{GliomaMRI} and \textbf{MammoBench} exhibit substantially weaker in-domain generalization, with Avg.~Source OUT scores well below this level, suggesting that the available training data is insufficient to even fit the source distribution well, let alone generalize to held-out domains.

\begin{table}[ht]
\centering
\caption{
    Balanced accuracy (\%) after pre-training on the \textbf{WILDSCamelyon} benchmark across five hospitals used as separate domains. Each row represents a hospital dataset which was used for training; per-hospital columns report performance on the respective \textit{IN} and \textit{OUT} splits. \textbf{Avg. Source} averages over training hospitals, \textbf{Avg. Target} over out-of-distribution dataset.
}
\label{tab:pre_wildscamelyon}
\begin{subtable}{\textwidth}
\centering
\caption{WILDSCamelyon -- ViT-B/16}
\label{tab:wilds_vit}
\resizebox{\textwidth}{!}{\begin{tabular}{lcccccccccc|cccc}
\toprule
Domain & \multicolumn{2}{c}{Hospital 1} & \multicolumn{2}{c}{Hospital 2} & \multicolumn{2}{c}{Hospital 3} & \multicolumn{2}{c}{Hospital 4} & \multicolumn{2}{c}{Hospital 5} & \multicolumn{2}{c}{Avg. Source} & \multicolumn{2}{c}{Avg. Target} \\
Split & IN & OUT & IN & OUT & IN & OUT & IN & OUT & IN & OUT & IN & OUT & IN & OUT \\
Training Domains (Source) &  &  &  &  &  &  &  &  &  &  &  &  &  &  \\
\midrule
Hospital 1 & 99.0 & 98.9 & 94.6 & 94.1 & 96.3 & 96.5 & 96.5 & 96.6 & 94.1 & 94.1 & 99.0 & 98.9 & 95.4 & 95.3 \\
Hospital 2 & 81.9 & 82.5 & 98.5 & 98.1 & 66.7 & 67.0 & 89.6 & 89.6 & 58.9 & 58.8 & 98.5 & 98.1 & 74.3 & 74.5 \\
Hospital 3 & 95.4 & 95.3 & 86.3 & 86.2 & 98.6 & 98.6 & 91.7 & 91.5 & 79.0 & 79.5 & 98.6 & 98.6 & 88.1 & 88.1 \\
Hospital 4 & 92.6 & 92.8 & 89.0 & 89.0 & 91.7 & 91.6 & 98.8 & 98.6 & 76.6 & 76.7 & 98.8 & 98.6 & 87.5 & 87.5 \\
Hospital 5 & 92.2 & 92.2 & 85.0 & 84.5 & 90.0 & 90.1 & 91.7 & 91.8 & 98.8 & 98.8 & 98.8 & 98.8 & 89.7 & 89.6 \\
\bottomrule
\end{tabular}
}
\end{subtable}
 
\vspace{0.4cm}
 
\begin{subtable}{\textwidth}
\centering
\caption{WILDSCamelyon -- ResNet50-GN}
\label{tab:wilds_resnet}
\resizebox{\textwidth}{!}{\begin{tabular}{lcccccccccc|cccc}
\toprule
Domain & \multicolumn{2}{c}{Hospital 1} & \multicolumn{2}{c}{Hospital 2} & \multicolumn{2}{c}{Hospital 3} & \multicolumn{2}{c}{Hospital 4} & \multicolumn{2}{c}{Hospital 5} & \multicolumn{2}{c}{Avg. Source} & \multicolumn{2}{c}{Avg. Target} \\
Split & IN & OUT & IN & OUT & IN & OUT & IN & OUT & IN & OUT & IN & OUT & IN & OUT \\
Training Domains (Source) &  &  &  &  &  &  &  &  &  &  &  &  &  &  \\
\midrule
Hospital 1 & 99.2 & 99.0 & 94.9 & 94.8 & 95.3 & 95.4 & 96.4 & 96.5 & 79.6 & 79.7 & 99.2 & 99.0 & 91.6 & 91.6 \\
Hospital 2 & 90.7 & 90.5 & 99.5 & 98.9 & 62.4 & 63.0 & 92.1 & 92.2 & 64.3 & 64.2 & 99.5 & 98.9 & 77.4 & 77.5 \\
Hospital 3 & 95.4 & 95.3 & 88.0 & 87.3 & 99.1 & 98.9 & 90.2 & 90.4 & 96.1 & 96.4 & 99.1 & 98.9 & 92.4 & 92.4 \\
Hospital 4 & 94.2 & 94.6 & 93.7 & 93.6 & 84.2 & 84.4 & 98.9 & 98.8 & 62.1 & 62.1 & 98.9 & 98.8 & 83.6 & 83.7 \\
Hospital 5 & 93.3 & 93.5 & 90.1 & 90.0 & 92.1 & 92.3 & 91.3 & 91.3 & 98.6 & 98.6 & 98.6 & 98.6 & 91.7 & 91.8 \\
\bottomrule
\end{tabular}
}
\end{subtable}
 
\end{table}

\begin{table}[ht]
\centering
  \caption{
    Balanced accuracy (\%) after pre-training on the \textbf{Histopantum} benchmark across four organ datasets used as separate domains. Each row represents the set of datasets which were used for training; per-dataset columns report performance on the respective \textit{IN} and \textit{OUT} splits. \textbf{Avg. Source} averages over training datasets, \textbf{Avg. Target} over out-of-distribution datasets.
}
\label{tab:pre_histopantum}
 
\begin{subtable}{\textwidth}
\centering
\caption{Histopantum -- ViT-B/16}
\label{tab:histo_vit}
\resizebox{\textwidth}{!}{\begin{tabular}{lcccccccc|cccc}
\toprule
Domain & \multicolumn{2}{c}{colon} & \multicolumn{2}{c}{ovarian} & \multicolumn{2}{c}{stomach} & \multicolumn{2}{c}{uterus} & \multicolumn{2}{c}{Avg. Source} & \multicolumn{2}{c}{Avg. Target} \\
Split & IN & OUT & IN & OUT & IN & OUT & IN & OUT & IN & OUT & IN & OUT \\
Training Domains (Source) &  &  &  &  &  &  &  &  &  &  &  &  \\
\midrule
colon & 98.8 & 98.2 & 92.3 & 91.9 & 86.1 & 85.3 & 84.0 & 84.1 & 98.8 & 98.2 & 87.5 & 87.1 \\
ovarian & 87.3 & 86.6 & 99.1 & 98.9 & 81.2 & 79.9 & 93.9 & 94.2 & 99.1 & 98.9 & 87.5 & 86.9 \\
stomach & 89.1 & 88.8 & 85.5 & 85.3 & 98.4 & 97.0 & 86.2 & 86.5 & 98.4 & 97.0 & 86.9 & 86.9 \\
uterus & 83.9 & 83.2 & 96.1 & 96.0 & 84.8 & 83.4 & 99.0 & 98.9 & 99.0 & 98.9 & 88.3 & 87.5 \\
\bottomrule
\end{tabular}
}
\end{subtable}
 
\vspace{0.4cm}
 
\begin{subtable}{\textwidth}
\centering
\caption{Histopantum -- ResNet50-GN}
\label{tab:histo_resnet}
\resizebox{\textwidth}{!}{\begin{tabular}{lcccccccc|cccc}
\toprule
Domain & \multicolumn{2}{c}{colon} & \multicolumn{2}{c}{ovarian} & \multicolumn{2}{c}{stomach} & \multicolumn{2}{c}{uterus} & \multicolumn{2}{c}{Avg. Source} & \multicolumn{2}{c}{Avg. Target} \\
Split & IN & OUT & IN & OUT & IN & OUT & IN & OUT & IN & OUT & IN & OUT \\
Training Domains (Source) &  &  &  &  &  &  &  &  &  &  &  &  \\
\midrule
colon & 99.3 & 98.5 & 91.3 & 91.6 & 83.9 & 82.8 & 85.1 & 85.4 & 99.3 & 98.5 & 86.8 & 86.6 \\
ovarian & 87.3 & 86.9 & 99.1 & 98.6 & 85.0 & 83.6 & 91.9 & 92.1 & 99.1 & 98.6 & 88.1 & 87.5 \\
stomach & 90.8 & 90.5 & 86.7 & 86.4 & 97.7 & 96.7 & 79.3 & 79.2 & 97.7 & 96.7 & 85.6 & 85.4 \\
uterus & 83.1 & 82.7 & 96.5 & 96.3 & 84.0 & 82.3 & 98.9 & 99.0 & 98.9 & 99.0 & 87.9 & 87.1 \\
\bottomrule
\end{tabular}
}
\end{subtable}
 
\end{table}

\begin{table}[ht]
\centering
 \caption{
    Balanced accuracy (\%) after pre-training on the \textbf{GliomaMRI} benchmark across three dataset used as separate domains. Each row represents the set of datasets which were used for training; per-dataset columns report performance on the respective \textit{IN} and \textit{OUT} splits. \textbf{Avg. Source} averages over training datasets, \textbf{Avg. Target} over out-of-distribution datasets.
}
\label{tab:pre_glioma}
\begin{subtable}{\textwidth}
\centering
\caption{GliomaMRI -- MONAI ViT-T/16}
\label{tab:glioma_vit}
\resizebox{\textwidth}{!}{\begin{tabular}{lcccccc|cccc}
\toprule
Domain & \multicolumn{2}{c}{erasmus} & \multicolumn{2}{c}{tcga} & \multicolumn{2}{c}{ucsf} & \multicolumn{2}{c}{Avg. Source} & \multicolumn{2}{c}{Avg. Target} \\
Split & IN & OUT & IN & OUT & IN & OUT & IN & OUT & IN & OUT \\
Training Domains (Source) &  &  &  &  &  &  &  &  &  &  \\
\midrule
erasmus, tcga & 88.7 & 47.4 & 98.5 & 45.4 & 43.6 & 33.0 & 93.6 & 46.4 & 43.6 & 33.0 \\
erasmus, ucsf & 100.0 & 50.8 & 37.6 & 41.1 & 100.0 & 33.1 & 100.0 & 42.0 & 37.6 & 41.1 \\
tcga, ucsf & 37.0 & 37.6 & 100.0 & 43.7 & 100.0 & 36.4 & 100.0 & 40.0 & 37.0 & 37.6 \\
\bottomrule
\end{tabular}
}
\end{subtable}
 
\vspace{0.4cm}
 
\begin{subtable}{\textwidth}
\centering
\caption{GliomaMRI -- MONAI ResNet10}
\label{tab:glioma_resnet}
\resizebox{\textwidth}{!}{\begin{tabular}{lcccccc|cccc}
\toprule
Domain & \multicolumn{2}{c}{erasmus} & \multicolumn{2}{c}{tcga} & \multicolumn{2}{c}{ucsf} & \multicolumn{2}{c}{Avg. Source} & \multicolumn{2}{c}{Avg. Target} \\
Split & IN & OUT & IN & OUT & IN & OUT & IN & OUT & IN & OUT \\
Training Domains (Source) &  &  &  &  &  &  &  &  &  &  \\
\midrule
erasmus, tcga & 100.0 & 64.0 & 100.0 & 40.7 & 54.4 & 41.1 & 100.0 & 52.4 & 54.4 & 41.1 \\
erasmus, ucsf & 100.0 & 55.2 & 42.9 & 41.7 & 100.0 & 44.9 & 100.0 & 50.0 & 42.9 & 41.7 \\
tcga, ucsf & 42.9 & 45.8 & 100.0 & 50.0 & 100.0 & 40.2 & 100.0 & 45.1 & 42.9 & 45.8 \\
\bottomrule
\end{tabular}
}
\end{subtable}

\end{table}

\begin{table}[ht]
\centering
 \caption{
    Balanced accuracy (\%) after pre-training on the \textbf{MammoBench} benchmark across five dataset used as separate domains. Each row represents the set of datasets which were used for training; per-dataset columns report performance on the respective \textit{IN} and \textit{OUT} splits. \textbf{Avg. Source} averages over training datasets, \textbf{Avg. Target} over out-of-distribution datasets.
}
\label{tab:pre_mammobench}
\begin{subtable}{\textwidth}
\centering
\caption{MammoBench -- ViT-B/16}
\label{tab:mammo_vit}
\resizebox{\textwidth}{!}{\begin{tabular}{lcccccccccc|cccc}
\toprule
Domain & \multicolumn{2}{c}{cdd-cesm} & \multicolumn{2}{c}{ddsm} & \multicolumn{2}{c}{dmid} & \multicolumn{2}{c}{inbreast} & \multicolumn{2}{c}{kau-bcmd} & \multicolumn{2}{c}{Avg. Source} & \multicolumn{2}{c}{Avg. Target} \\
Split & IN & OUT & IN & OUT & IN & OUT & IN & OUT & IN & OUT & IN & OUT & IN & OUT \\
Training Domains (Source) &  &  &  &  &  &  &  &  &  &  &  &  &  &  \\
\midrule
cdd-cesm, ddsm, dmid, inbreast & 90.8 & 66.5 & 81.3 & 75.0 & 100.0 & 80.4 & 99.8 & 61.8 & 48.5 & 48.2 & 93.0 & 70.9 & 48.5 & 48.2 \\
cdd-cesm, ddsm, dmid, kau-bcmd & 90.0 & 71.7 & 74.3 & 72.4 & 99.7 & 75.1 & 60.7 & 65.0 & 62.3 & 61.8 & 81.6 & 70.2 & 60.7 & 65.0 \\
cdd-cesm, ddsm, inbreast, kau-bcmd & 82.6 & 60.1 & 73.8 & 70.9 & 52.7 & 52.4 & 99.8 & 50.9 & 72.4 & 68.4 & 82.1 & 62.6 & 52.7 & 52.4 \\
cdd-cesm, dmid, inbreast, kau-bcmd & 90.4 & 67.6 & 56.9 & 54.5 & 99.7 & 77.1 & 100.0 & 52.5 & 82.5 & 79.5 & 93.2 & 69.2 & 56.9 & 54.5 \\
ddsm, dmid, inbreast, kau-bcmd & 53.8 & 52.0 & 77.9 & 69.9 & 99.7 & 75.9 & 100.0 & 59.5 & 59.9 & 57.8 & 84.4 & 65.8 & 53.8 & 52.0 \\
\bottomrule
\end{tabular}
}
\end{subtable}
 
\vspace{0.4cm}
 
\begin{subtable}{\textwidth}
\centering
\caption{MammoBench -- ResNet50-GN}
\label{tab:mammo_resnet}
\resizebox{\textwidth}{!}{\begin{tabular}{lcccccccccc|cccc}
\toprule
Domain & \multicolumn{2}{c}{cdd-cesm} & \multicolumn{2}{c}{ddsm} & \multicolumn{2}{c}{dmid} & \multicolumn{2}{c}{inbreast} & \multicolumn{2}{c}{kau-bcmd} & \multicolumn{2}{c}{Avg. Source} & \multicolumn{2}{c}{Avg. Target} \\
Split & IN & OUT & IN & OUT & IN & OUT & IN & OUT & IN & OUT & IN & OUT & IN & OUT \\
Training Domains (Source) &  &  &  &  &  &  &  &  &  &  &  &  &  &  \\
\midrule
cdd-cesm, ddsm, dmid, inbreast & 86.0 & 73.7 & 79.0 & 76.1 & 100.0 & 82.0 & 100.0 & 58.6 & 50.1 & 51.0 & 91.2 & 72.6 & 50.1 & 51.0 \\
cdd-cesm, ddsm, dmid, kau-bcmd & 90.2 & 68.2 & 76.5 & 72.3 & 100.0 & 78.8 & 59.5 & 65.6 & 74.4 & 76.3 & 85.3 & 73.9 & 59.5 & 65.6 \\
cdd-cesm, ddsm, inbreast, kau-bcmd & 91.6 & 74.1 & 76.4 & 73.5 & 63.7 & 61.6 & 100.0 & 56.2 & 77.0 & 71.4 & 86.2 & 68.8 & 63.7 & 61.6 \\
cdd-cesm, dmid, inbreast, kau-bcmd & 80.6 & 76.8 & 54.6 & 53.7 & 88.7 & 81.6 & 94.6 & 57.1 & 90.5 & 88.8 & 88.6 & 76.1 & 54.6 & 53.7 \\
ddsm, dmid, inbreast, kau-bcmd & 62.8 & 65.2 & 78.1 & 76.0 & 97.9 & 77.5 & 98.8 & 59.5 & 79.9 & 78.3 & 88.7 & 72.8 & 62.8 & 65.2 \\
\bottomrule
\end{tabular}
}
\end{subtable}
\end{table}

\newpage
\section{Compute Resources}\label{sec_app:compute}
For our experiments we utilize a SLURM cluster with varying hardware. GPUs used include NVIDIA RTX A5000 24GB, A100 40GB, H100 40GB, and H100 80GB.
Except for the experiments on the 3D MRI dataset GliomaMRI, all experiments can be reproduced on a GPU with at least 24GB of VRAM. For pre-training the GliomaMRI models, we required the H100 80GB GPUs as the 3D volumes and models consumed more than 50GB of VRAM.

\begin{table}[ht]
    \centering
    \caption{Average adaptation runtime in seconds per adaptation source--target configuration
    for medical datasets. \textbf{Bold} = lowest runtime per column; \underline{underline} =
    second lowest. DSBR consistently ranks among the fastest methods across both
    architectures, matching the baseline with the lowest runtime in nearly every setting.}
    \label{tab:runtimes_medical}
    \resizebox{\textwidth}{!} {
    \begin{tabular}{llcccc}
    \toprule
     Model Type & \textbf{Method} & \textbf{GliomaMRI} & \textbf{Histopantum} & \textbf{MammoBench} & \textbf{WILDSCamelyon} \\
    \midrule
    \multirow[t]{5}{*}{ResNet}
     & Tent~\cite{wang2021tent} & \underline{22}          & \textbf{82}         & \underline{63} & \textbf{149} \\
     & SAR~\cite{niu2023towards}  & 26                  & 100                 & 64         & 209 \\
     & ROID~\cite{marsden2024universal} & \textbf{21}          & 106                 & 65         & 207 \\
     & DeYO~\cite{lee2024entropy} & n/a          & \underline{85}          & \textbf{61}& \underline{167} \\
     \rowcolor{DSBR_purple!20} & \textbf{\textit{DSBR}}  & 23                  & \textbf{82}         & \underline{63} & \textbf{149} \\
    \midrule
    \multirow[t]{5}{*}{ViT}
     & Tent~\cite{wang2021tent} & 20         & \textbf{162}        & \textbf{70}& \textbf{395} \\
     & SAR~\cite{niu2023towards}  & \underline{19}          & 236                 & 79         & 613 \\
     & ROID~\cite{marsden2024universal} & \textbf{18}         & 241                 & 75         & 551 \\
     & DeYO~\cite{lee2024entropy} & n/a               & 184                 & \underline{74} & 426 \\
     \rowcolor{DSBR_purple!20} & \textbf{\textit{DSBR}} & \underline{19}          & \underline{167}         & \textbf{70}& \underline{396} \\
    \bottomrule
    \end{tabular}
    }
\end{table}

\begin{table}[ht]
    \centering
    \caption{Average adaptation runtime in seconds per adaptation domain (corruption) for all
    ImageNet-C based experiments. \textbf{Bold} = lowest runtime per column; \underline{underline}
    = second lowest. DSBR achieves the lowest or second-lowest runtime in the majority of
    settings, making it a computationally efficient method while remaining competitive
    in accuracy.}
    \label{tab:runtimes_imagenetc}
    \resizebox{\textwidth}{!} {
    \begin{tabular}{llcccc}
    \toprule
     Model Type & \textbf{Method} & \textbf{Batch Size 1} & \textbf{Default} & \textbf{Label Shifts} & \textbf{Mixed Corruptions} \\
    \midrule
    \multirow[t]{6}{*}{ResNet}
     & Tent~\cite{wang2021tent}                  & 737  & \underline{77}          & \textbf{111}& 876 \\
     & SAR~\cite{niu2023towards}                   & 1349         & 101                 & 175         & 1418 \\
     & ROID~\cite{marsden2024universal}                  & 1632         & 98                  & 166         & 1173 \\
     & DeYO~\cite{lee2024entropy}                  & \textbf{692} & 81                  & 131         & 896 \\
     \rowcolor{DSBR_purple!20} & \textbf{\textit{DSBR}}($\alpha=0.99$)   & 734          & 82                  & 117         & \textbf{802} \\
     \rowcolor{DSBR_purple!20} & \textbf{\textit{DSBR}}($\alpha=0.9$)    & \underline{717}  & \textbf{75}         & \underline{115} & \underline{817} \\
    \midrule
    \multirow[t]{6}{*}{ViT}
     & Tent~\cite{wang2021tent}                  & \textbf{555} & 186                 & \textbf{428}& \textbf{2967} \\
     & SAR~\cite{niu2023towards}                   & 1142         & 241                 & 808         & 6118 \\
     & ROID~\cite{marsden2024universal}                  & 1273         & 206                 & 526         & 4456 \\
     & DeYO~\cite{lee2024entropy}                  & 686          & \underline{182}         & 505         & 4058 \\
     \rowcolor{DSBR_purple!20} & \textbf{\textit{DSBR}}($\alpha=0.99$)   & \underline{612}  & \textbf{167}        & 478         & \underline{3162} \\
     \rowcolor{DSBR_purple!20} & \textbf{\textit{DSBR}}($\alpha=0.9$)    & 639          & \underline{172}         & \underline{477} & 3362 \\
    \bottomrule
    \end{tabular}
    }
\end{table}

Even though we utilize different GPUs in our adaptation experiments, the total consumed time per experiment is almost equal for all GPUs.
\cref{tab:runtimes_medical} and \ref{tab:runtimes_imagenetc} report the average runtime of each method for every dataset. Note, that these numbers refer to a single adaptation run of a single configuration and we repeat each experiment for 5 and 3 seeds for medical datasets and ImageNet-C respectively.
For example, WILDSCamelyon has a total number of 20 source--target configurations, so the total runtime of each method needs to be multiplied by $20 (\text{configurations}) \cdot 5 (\text{seeds}) = 100$.

To pretrain all models using the codebase build on Domainbed~\cite{gulrajani2021in}, we require approximately 10 hours for each model on a dataset.

\section{High Gradient Norm is an Effect of Model Collapse rather than its Cause}\label{sec:grad_norm}
\citet{niu2023towards} attribute model collapse to a small subset of samples that produce disproportionately large gradient updates and destabilize adaptation.
Their argument is empirical: during runs that end in collapse, the per-batch gradient norm grows sharply just before predictions concentrate on a single class.
The proposed remedy is to identify and discard high-gradient-norm samples before they contribute to the update.

We test this attribution directly.
For each adaptation configuration we precompute every sample's \emph{initial gradient norm} by running the model on it as a single-sample batch and recording the resulting gradient before any update is applied.
This gives us a per-sample ranking of the gradient signal each input would contribute to the very first adaptation step.
We then form subsets of the data corresponding to fixed percentiles of this ranking and run full adaptation on each subset separately.
If the high-gradient-norm hypothesis is correct, subsets containing the highest-percentile samples should produce visibly worse outcomes than subsets containing only low-percentile samples.

\begin{figure}
    \centering
    \includegraphics[width=\linewidth]{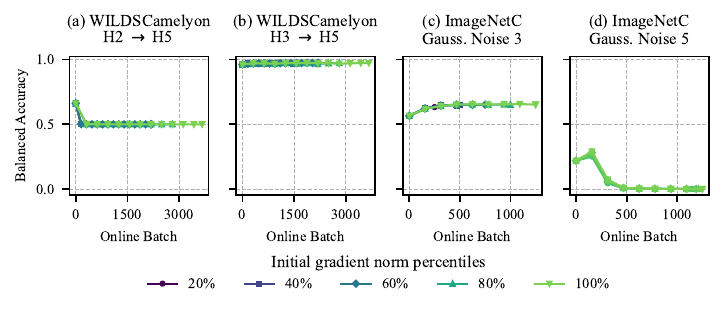}
    \caption{Initial gradient norm has no effect on entropy minimization. We compute the initial gradient norm for each sample by running adaptation on a pre-trained ResNet50-GN with batch size 1 and resetting the model after each step. Afterward, we run normal adaptation on varying percentiles of gradient norms.   }
    \label{fig:filter_grad}
\end{figure}

\cref{fig:filter_grad} shows the result.
Performance is essentially equal across gradient-norm percentiles, both in configurations where adaptation succeeds (b, c) and in configurations where it fails (a, d).
The samples with the largest initial gradient norms are not, on their own, harmful to adaptation: removing them does not prevent collapse in the failure cases, and including them does not damage adaptation in the success cases.

We therefore argue that the temporal correlation observed by \citet{niu2023towards} is read in the wrong direction.
The high gradient norms that precede collapse are an effect of the model already drifting toward a degenerate state, in which the loss surface becomes locally steep.
Whatever drives collapse must therefore originate upstream of the gradient signal, in the geometry of the feature space.
We note that the SAR algorithm proposed by \citet{niu2023towards} does not filter on gradient norms but on entropy in practice, motivated as a model- and dataset-independent proxy for gradient-norm filtering; we state in \cref{sec_app:entropy} that entropy filtering does work, but for a reason different from the one this proxy implies.

\section{Ability to Recover Rare Classes and Calibrate Uncertainty}

\textbf{Recovery of rare classes at test-time.} \browntext{As described in methodology, DSBR up-weights each sample's contribution to the entropy loss through inverse-frequency reweighting. Through the EMA estimate of cluster frequencies, this reweighting also persists across batches in which a class is absent. The predicted distribution therefore stays spread across classes rather than collapsing to one, allowing rare classes suppressed by the domain shift to re-emerge in DSBR's predictions.}

\browntext{To further highlight this behavior, which we term class recovery, we designed an additional experiment in~\cref{tab:class_recovery}. Class recovery refers to the reappearance, after adaptation, of classes that the source model rarely or never predicts under domain shift, with their predicted frequencies moving closer to their true frequencies. The experiment also extends the analysis in~\cref{tab:pred_bias} of the main paper. For ImageNet-C, we track the classes with the lowest prediction probability before adaptation and measure whether Tent or DSBR are able to recover them. DSBR's predictions are closer to the true class frequency in 8 of 10 cases, Tent is closer in one case, and the remaining case is a tie. This shows that DSBR is able to recover suppressed classes more reliably.}

\begin{table}[h]
\centering
\caption{\purpletext{\textbf{Recovery of rare classes at test-time}  on ImageNet-C with ResNet50-GN.} \purpletext{Predicted class distributions before and after Tent (Entropy Minimization) and DSBR (ours). DSBR recovers the rare classes more often than Tent.}}
\label{tab:class_recovery}
\begin{tabular}{lccccc}
\toprule
\textbf{Class from ImageNet} & \textbf{True} & \textbf{Before} & {\setlength{\fboxsep}{1pt}\colorbox{DSBR_purple!20}{\textbf{DSBR}}} & \textbf{Tent} & \begin{tabular}[c]{@{}c@{}}\textbf{Method closer}\\\textbf{to True (Best)}\end{tabular} \\
\midrule
African elephant & 0.09\% & 0.00\% & \textbf{0.04\%} & 0.01\% & DSBR \\
Backpack         & 0.06\% & 0.00\% & \textbf{0.03\%} & 0.01\% & DSBR \\
Breakwater       & 0.08\% & 0.00\% & \textbf{0.05\%} & 0.01\% & DSBR \\
Chocolate sauce  & 0.11\% & 0.00\% & 0.16\% & \textbf{0.09\%} & Tent \\
Garden spider    & 0.16\% & 0.00\% & \textbf{0.15\%} & 0.10\% & DSBR \\
Groom            & 0.04\% & 0.00\% & \textbf{0.04\%} & 0.02\% & DSBR \\
Median (other)   & 0.10\% & 0.08\% & \textbf{0.10\%} & 0.09\% & DSBR \\
Pizza            & 0.11\% & 0.01\% & \textbf{0.06\%} & 0.02\% & DSBR \\
Typewriter kbd   & 0.07\% & 0.01\% & \textbf{0.03\%} & 0.01\% & DSBR \\
Unicycle         & 0.05\% & 0.01\% & \textbf{0.06\%} & \textbf{0.06\%} & Both \\
Wolf spider      & 0.11\% & 0.00\% & \textbf{0.14\%} & 0.05\% & DSBR \\
\bottomrule
\end{tabular}
\end{table}

\textbf{Addressing uncertainty.} \purpletext{To demonstrate the ability to handle uncertain scenarios, \cref{tab:medical_ece} also reports the Expected Calibration Error (ECE)~\cite{guo2017calibration} for all four medical datasets on both backbones, where lower ECE is better. With ResNet, DSBR achieves the best average ECE change among all TTA methods. It also ranks second on Histopathology and WILDSCamelyon, where it is the only method that reduces ECE on both. With ViT, SAR and Tent obtain the lowest average ECE change, while DSBR increases ECE only moderately, by less than ROID, whose calibration is unstable on GliomaMRI. Notably, DSBR relies solely on the proposed reweighting of the entropy loss, without the auxiliary tasks or models, smoothing, weight ensembling, or augmentation-based consistency used by other baselines. Such components are orthogonal to our method.}
\begin{table}[h]
\centering
\newcommand{\dpos}[1]{\,{\scriptsize #1}}
\newcommand{\dneg}[1]{\,{\scriptsize $-$#1}}
\newcommand{\dneu}[1]{\,{\scriptsize #1}}
\caption{\purpletext{\textbf{Addressing uncertainty} with ResNet and ViT backbones using ECE as the uncertainty metric (lower is better). The subscripts show the change relative to the source model, where a negative value indicates a decrease in ECE and therefore an improvement.}}
\label{tab:medical_ece}
\resizebox{1.0\linewidth}{!}{
\begin{tabular}{lccccc}
\toprule
Model \& Method & \textbf{Histopantum} & \textbf{MammoBench} & \textbf{WILDSCamelyon} & \textbf{GliomaMRI} & \textbf{$\Delta$} \textbf{Average} \\
\midrule
ResNet & 9.9\dneu{0.0}$_{\pm0.0}$ & 38.4\dneu{0.0}$_{\pm0.0}$ & 8.4\dneu{0.0}$_{\pm0.0}$ & 16.4\dneu{0.0}$_{\pm0.0}$ & \dneu{0.0}$_{\pm0.0}$ \\
~~$\bullet~$Tent~\cite{wang2021tent} & \textbf{8.5\dneg{1.4}$_{\pm0.1}$} & \underline{39.0\dpos{0.6}$_{\pm0.1}$} & 9.9\dpos{1.5}$_{\pm0.0}$ & \underline{19.1\dpos{2.7}$_{\pm1.3}$} & \dpos{0.9}$_{\pm0.4}$ \\
~~$\bullet~$SAR~\cite{lee2023towards} & 9.9\dneu{0.0}$_{\pm0.0}$ & \textbf{38.4\dneu{0.0}$_{\pm0.0}$} & 8.4\dneu{0.0}$_{\pm0.0}$ & 19.2\dpos{2.8}$_{\pm1.1}$ & \underline{\dpos{0.7}$_{\pm0.3}$} \\
~~$\bullet~$ROID~\cite{marsden2024universal} & 11.4\dpos{1.5}$_{\pm1.3}$ & 46.0\dpos{7.6}$_{\pm2.8}$ & \textbf{2.6\dneg{5.7}$_{\pm0.3}$} & \textbf{18.8\dpos{2.4}$_{\pm1.2}$} & \dpos{1.4}$_{\pm1.4}$ \\
~~$\bullet~$DeYO~\cite{lee2024entropy} & 22.3\dpos{12.4}$_{\pm3.1}$ & 44.8\dpos{6.4}$_{\pm0.5}$ & 5.0\dneg{3.3}$_{\pm0.3}$ & n/a & \dpos{5.2}$_{\pm1.3}$ \\
\rowcolor{DSBR_purple!20}~~$\bullet~$\textbf{\textit{DSBR}}($\alpha=0.9$) & \underline{9.6\dneg{0.3}$_{\pm0.5}$} & 41.7\dpos{3.2}$_{\pm0.8}$ & \underline{2.9\dneg{5.5}$_{\pm0.1}$} & 19.6\dpos{3.2}$_{\pm1.7}$ & \textbf{\dpos{0.2}$_{\pm0.8}$} \\
\midrule
ViT & 8.6\dneu{0.0}$_{\pm0.0}$ & 41.5\dneu{0.0}$_{\pm0.0}$ & 7.3\dneu{0.0}$_{\pm0.0}$ & 34.7\dneu{0.0}$_{\pm0.0}$ & \dneu{0.0}$_{\pm0.0}$ \\
~~$\bullet~$Tent~\cite{wang2021tent} & \textbf{8.7\dpos{0.1}$_{\pm0.0}$} & \textbf{41.5\dneu{0.0}$_{\pm0.2}$} & 8.8\dpos{1.5}$_{\pm0.2}$ & \textbf{34.6\dneg{0.1}$_{\pm1.2}$} & \underline{\dpos{0.4}$_{\pm0.4}$} \\
~~$\bullet~$SAR~\cite{lee2023towards} & \textbf{8.7\dneu{0.0}$_{\pm0.4}$} & \textbf{41.5\dneu{0.0}$_{\pm0.0}$} & 7.3\dneu{0.0}$_{\pm0.0}$ & \underline{34.7\dneu{0.0}$_{\pm0.0}$} & \textbf{\dneu{0.0}$_{\pm0.1}$} \\
~~$\bullet~$ROID~\cite{marsden2024universal} & \underline{8.8\dpos{0.2}$_{\pm0.3}$} & \underline{42.7\dpos{1.2}$_{\pm0.7}$} & \textbf{2.0\dneg{5.3}$_{\pm0.2}$} & 45.8\dpos{11.1}$_{\pm8.8}$ & \dpos{1.8}$_{\pm2.5}$ \\
~~$\bullet~$DeYO~\cite{lee2024entropy} & 9.1\dpos{0.4}$_{\pm0.5}$ & 43.1\dpos{1.6}$_{\pm0.2}$ & \underline{3.8\dneg{3.5}$_{\pm1.9}$} & n/a & \dneg{0.5}$_{\pm0.9}$ \\
\rowcolor{DSBR_purple!20}~~$\bullet~$\textbf{\textit{DSBR}}($\alpha=0.9$) & 11.3\dpos{2.7}$_{\pm0.6}$ & 44.5\dpos{3.1}$_{\pm0.4}$ & 4.3\dneg{3.0}$_{\pm0.3}$ & 36.6\dpos{1.9}$_{\pm0.6}$ & \dpos{1.2}$_{\pm0.5}$ \\
\bottomrule
\end{tabular}
}
\end{table}

\begin{table}[t]
\centering
\scriptsize
\setlength{\tabcolsep}{1.5pt}
\resizebox{\textwidth}{!}{%
\begin{tabular}{l|ccc|cccc|cccc|cccc|c}
\toprule
 & \multicolumn{3}{c|}{Noise} & \multicolumn{4}{c|}{Blur} & \multicolumn{4}{c|}{Weather} & \multicolumn{4}{c|}{Digital} &  \\
 & Gauss. & Shot & Impul. & Defoc. & Glass & Motion & Zoom & Snow & Frost & Fog & Brit. & Contr. & Elastic & Pixel & JPEG & Avg. \\
\midrule
ResNet & 22.1$_{\pm0.0}$ & 23.0$_{\pm0.0}$ & 22.0$_{\pm0.0}$ & \underline{19.8$_{\pm0.0}$} & 11.4$_{\pm0.0}$ & 21.5$_{\pm0.0}$ & 25.0$_{\pm0.0}$ & 40.2$_{\pm0.0}$ & \textbf{46.9$_{\pm0.0}$} & 34.0$_{\pm0.0}$ & \textbf{68.8$_{\pm0.0}$} & 36.2$_{\pm0.0}$ & 18.5$_{\pm0.0}$ & 29.3$_{\pm0.0}$ & 52.6$_{\pm0.0}$ & 31.4$_{\pm0.0}$ \\
$\bullet$ Tent & 24.0$_{\pm1.5}$ & 22.9$_{\pm0.8}$ & 23.0$_{\pm1.6}$ & 19.1$_{\pm1.5}$ & 11.1$_{\pm3.5}$ & 21.7$_{\pm2.7}$ & 25.5$_{\pm1.6}$ & 37.7$_{\pm4.2}$ & \underline{46.2$_{\pm1.4}$} & 25.3$_{\pm4.5}$ & 67.9$_{\pm0.3}$ & 37.0$_{\pm5.3}$ & 17.5$_{\pm3.5}$ & 33.8$_{\pm0.3}$ & 52.9$_{\pm10.6}$ & 31.1$_{\pm2.9}$ \\
$\bullet$ SAR & 24.1$_{\pm1.6}$ & 23.4$_{\pm1.3}$ & 23.4$_{\pm1.4}$ & \textbf{20.3$_{\pm1.9}$} & 11.4$_{\pm3.8}$ & 22.0$_{\pm3.3}$ & 25.6$_{\pm1.6}$ & 37.9$_{\pm3.7}$ & 45.7$_{\pm1.6}$ & 29.4$_{\pm8.4}$ & 68.0$_{\pm0.4}$ & 36.8$_{\pm4.6}$ & 17.7$_{\pm3.6}$ & 32.5$_{\pm1.5}$ & 52.7$_{\pm10.4}$ & 31.4$_{\pm3.3}$ \\
$\bullet$ ROID & \textbf{31.8$_{\pm0.8}$} & \textbf{31.1$_{\pm2.0}$} & \textbf{31.3$_{\pm2.0}$} & 17.8$_{\pm1.4}$ & \textbf{16.3$_{\pm3.2}$} & \textbf{27.3$_{\pm4.2}$} & \underline{29.4$_{\pm2.9}$} & \textbf{42.4$_{\pm3.3}$} & 45.1$_{\pm0.9}$ & \textbf{46.1$_{\pm2.9}$} & 68.5$_{\pm0.2}$ & \textbf{42.3$_{\pm6.2}$} & \textbf{27.0$_{\pm3.8}$} & \textbf{43.4$_{\pm1.4}$} & 53.5$_{\pm9.8}$ & \textbf{36.9$_{\pm3.0}$} \\
$\bullet$ DeYO & \underline{28.8$_{\pm0.9}$} & 27.8$_{\pm1.7}$ & \underline{28.6$_{\pm3.7}$} & 18.5$_{\pm1.6}$ & 13.8$_{\pm4.0}$ & 25.4$_{\pm4.9}$ & 27.1$_{\pm2.3}$ & 38.9$_{\pm4.9}$ & 44.1$_{\pm1.9}$ & 22.1$_{\pm19.2}$ & \underline{68.6$_{\pm0.6}$} & 39.7$_{\pm5.8}$ & 21.1$_{\pm4.2}$ & \underline{42.9$_{\pm2.0}$} & 53.9$_{\pm10.1}$ & 33.4$_{\pm4.5}$ \\
\rowcolor{DSBR_purple!20}$\bullet$ \textbf{\textit{DSBR}} ($\alpha{=}0.99$) & 27.8$_{\pm1.3}$ & 27.8$_{\pm0.5}$ & 27.2$_{\pm1.6}$ & 18.1$_{\pm1.6}$ & 14.8$_{\pm3.4}$ & 26.4$_{\pm4.4}$ & 29.2$_{\pm2.5}$ & 40.1$_{\pm3.5}$ & 45.1$_{\pm1.4}$ & 33.9$_{\pm2.9}$ & 68.6$_{\pm0.3}$ & \underline{39.7$_{\pm6.2}$} & 24.3$_{\pm3.9}$ & 39.6$_{\pm0.6}$ & \textbf{54.7$_{\pm10.2}$} & 34.5$_{\pm3.0}$ \\
\rowcolor{DSBR_purple!20}$\bullet$ \textbf{\textit{DSBR}} ($\alpha{=}0.9$) & 28.2$_{\pm1.3}$ & \underline{28.1$_{\pm0.5}$} & 27.6$_{\pm1.2}$ & 18.4$_{\pm1.4}$ & \underline{15.3$_{\pm3.9}$} & \underline{26.8$_{\pm4.1}$} & \textbf{29.4$_{\pm2.4}$} & \underline{40.4$_{\pm3.1}$} & 45.2$_{\pm1.4}$ & \underline{35.0$_{\pm3.1}$} & 68.4$_{\pm0.4}$ & 39.7$_{\pm6.3}$ & \underline{24.7$_{\pm3.8}$} & 39.6$_{\pm0.5}$ & \underline{54.5$_{\pm10.0}$} & \underline{34.8$_{\pm2.9}$} \\
\midrule
ViT & 35.1$_{\pm0.0}$ & 32.1$_{\pm0.0}$ & 35.8$_{\pm0.0}$ & 31.4$_{\pm0.0}$ & 25.3$_{\pm0.0}$ & 39.5$_{\pm0.0}$ & 31.5$_{\pm0.0}$ & 24.5$_{\pm0.0}$ & 30.1$_{\pm0.0}$ & 54.7$_{\pm0.0}$ & 64.5$_{\pm0.0}$ & 49.0$_{\pm0.0}$ & 34.2$_{\pm0.0}$ & 53.2$_{\pm0.0}$ & 56.5$_{\pm0.0}$ & 39.8$_{\pm0.0}$ \\
$\bullet$ Tent & 45.1$_{\pm2.3}$ & 38.3$_{\pm2.2}$ & 45.1$_{\pm0.8}$ & 37.6$_{\pm2.8}$ & 33.8$_{\pm5.0}$ & 45.6$_{\pm4.6}$ & 37.5$_{\pm4.5}$ & 28.6$_{\pm4.7}$ & 41.2$_{\pm0.5}$ & 55.8$_{\pm1.7}$ & 67.2$_{\pm1.6}$ & 57.2$_{\pm3.3}$ & 37.6$_{\pm3.0}$ & 56.2$_{\pm2.4}$ & 59.1$_{\pm9.7}$ & 45.7$_{\pm3.3}$ \\
$\bullet$ SAR & 44.7$_{\pm1.8}$ & 38.4$_{\pm2.1}$ & 45.0$_{\pm0.7}$ & 39.3$_{\pm2.2}$ & 33.9$_{\pm4.2}$ & 45.2$_{\pm5.0}$ & 37.2$_{\pm4.8}$ & 27.0$_{\pm4.3}$ & 40.5$_{\pm1.7}$ & 56.1$_{\pm0.6}$ & 66.3$_{\pm2.5}$ & 56.2$_{\pm3.6}$ & 37.2$_{\pm2.7}$ & 55.3$_{\pm2.3}$ & 58.6$_{\pm9.4}$ & 45.4$_{\pm3.2}$ \\
$\bullet$ ROID & \underline{53.4$_{\pm1.0}$} & \textbf{50.1$_{\pm1.4}$} & \underline{53.5$_{\pm1.3}$} & \textbf{49.0$_{\pm4.5}$} & \textbf{46.6$_{\pm3.7}$} & \textbf{54.0$_{\pm4.0}$} & \textbf{48.3$_{\pm5.6}$} & \textbf{51.6$_{\pm3.1}$} & \textbf{54.2$_{\pm2.7}$} & \textbf{66.3$_{\pm0.6}$} & \underline{70.4$_{\pm1.3}$} & \textbf{64.2$_{\pm2.6}$} & \textbf{54.9$_{\pm3.0}$} & \textbf{61.9$_{\pm1.5}$} & \underline{62.9$_{\pm8.7}$} & \textbf{56.1$_{\pm3.0}$} \\
$\bullet$ DeYO & 50.9$_{\pm2.1}$ & 47.4$_{\pm2.2}$ & 51.9$_{\pm1.2}$ & 46.4$_{\pm4.8}$ & 41.9$_{\pm3.8}$ & 51.1$_{\pm6.1}$ & 42.2$_{\pm6.5}$ & \underline{41.3$_{\pm3.9}$} & \underline{51.0$_{\pm0.6}$} & 63.0$_{\pm1.4}$ & \textbf{70.8$_{\pm2.0}$} & 61.6$_{\pm4.0}$ & 47.1$_{\pm2.7}$ & 59.2$_{\pm2.5}$ & 60.8$_{\pm8.5}$ & 52.4$_{\pm3.5}$ \\
\rowcolor{DSBR_purple!20}$\bullet$ \textbf{\textit{DSBR}} ($\alpha{=}0.99$) & \textbf{53.5$_{\pm2.2}$} & \underline{49.2$_{\pm1.4}$} & \textbf{53.7$_{\pm1.0}$} & 46.0$_{\pm5.2}$ & 45.0$_{\pm4.1}$ & 53.7$_{\pm2.5}$ & 48.1$_{\pm3.6}$ & 38.6$_{\pm4.8}$ & 47.6$_{\pm2.5}$ & \underline{64.7$_{\pm1.5}$} & 69.5$_{\pm1.4}$ & \underline{63.0$_{\pm2.0}$} & 52.7$_{\pm2.0}$ & \underline{61.4$_{\pm0.8}$} & \textbf{62.9$_{\pm9.8}$} & \underline{54.0$_{\pm3.0}$} \\
\rowcolor{DSBR_purple!20}$\bullet$ \textbf{\textit{DSBR}} ($\alpha{=}0.9$) & 53.1$_{\pm1.7}$ & 48.5$_{\pm2.0}$ & 53.4$_{\pm1.1}$ & \underline{46.5$_{\pm3.9}$} & \underline{45.2$_{\pm3.8}$} & \underline{53.8$_{\pm3.0}$} & \underline{48.2$_{\pm3.6}$} & 39.2$_{\pm4.6}$ & 47.2$_{\pm1.8}$ & 64.3$_{\pm1.4}$ & 69.0$_{\pm2.1}$ & 62.7$_{\pm2.2}$ & \underline{53.4$_{\pm4.1}$} & 60.8$_{\pm2.3}$ & 62.8$_{\pm9.7}$ & 53.9$_{\pm3.1}$ \\
\bottomrule
\end{tabular}}
\caption{\textbf{Severe class imbalance} on ImageNet-C, following the imbalanced protocol of \citet{weerasinghe2026logit} for ResNet and ViT backbones. }
\label{tab:imbalance}
\end{table}

\textbf{Severe class imbalance.} \purpletext{Beyond the datasets in the main paper, which already exhibit the class imbalance typical of the medical domain, we also evaluate under the imbalanced protocol of \citet{weerasinghe2026logit} (Table 1). We apply it to all ImageNet-C corruptions on both backbones and report balanced accuracy (BA), which we provide in ~\cref{tab:imbalance}. With ViT, DSBR improves average BA over the source model by $+14.2$ ($54.0 \pm 3.0$), on par with the best method, ROID~\cite{marsden2024universal} ($+16.3$; $56.1 \pm 3.0$), within error margins, and well above Tent~\cite{wang2021tent} ($+5.9$; $45.7 \pm 3.3$). ResNet leads to a similar conclusion. Together with the wild TTA scenarios in~\cref{tab:imagenetc_results} of the main paper, this indicates that DSBR adapts stably under severe imbalance, whereas methods such as ROID collapse under batch size 1 and label shift on the ResNet backbone.}

\newpage
\section{Entropy filtering as one solution for cluster separation}
\label{sec_app:entropy}
\begin{wrapfigure}{r}{0.4\linewidth} %
  \vspace{-20pt}
  \centering
  \includegraphics[width=\linewidth]{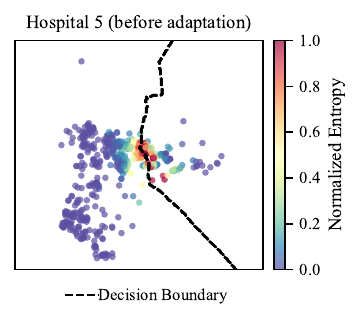}
  \\[-1.0em]
    \caption{Entropy peaks near the decision boundary with the setup as in \cref{fig:domain_shift_bias}, samples are colored by prediction entropy normalized by $\log n$, with $n=2$.}
    \label{fig:entropy}
\end{wrapfigure}

The cluster-refinement view also explains why entropy filtering, despite our argument in Section~\ref{sec:grad_norm} that the gradient-norm motivation for it is false, can still work.
Samples close to the decision boundary have high entropy, since the model's confidence falls off with distance from the boundary, as illustrated in \cref{fig:entropy}.
Under cluster merging, the region near the boundary is densely populated with samples that have been incorrectly reassigned across it.
Filtering high-entropy samples \citep{niu2022efficient, niu2023towards, lee2024entropy} therefore removes a disproportionate share of these reassigned samples from the gradient update, and additionally cuts an artificial trench through the boundary region that helps the merged clusters separate again.
This is, in our view, the actual mechanism by which entropy-based filtering stabilizes Tent, rather than the gradient-norm proxy argument given in \citet{niu2023towards}.
We do not pursue filtering in our work: it is wasteful when data are scarce, and the samples it discards are possibly very informative about how the model is misaligned with the target domain.
The next section develops an alternative that addresses prediction bias directly while keeping every sample in the adaptation signal.

\begin{table}[t]
    \vspace{-2mm}
    \centering
    \caption{ROC-AUC (\%) on the four medical benchmarks, in the same configuration as \cref{tab:results_medical}.$^*$Disabled \emph{consistency} component because it relies on 2D augmentations. $^\dagger$Not available because whole method relies on 2D augmentations. }
    \label{tab:results_medical_roc_auc}
    \resizebox{1.0\linewidth}{!}{
        \begin{tabular}{lccccc}
        \toprule
        Model \& Method & Histopantum & MammoBench & WILDSCamelyon & GliomaMRI & $\Delta$ Average \\
        \midrule
        ResNet & 94.1$_{\pm0.0}$ & 61.3$_{\pm0.0}$ & 97.0$_{\pm0.0}$ & \textbf{70.0$_{\pm0.0}$} & \underline{\neu0.0$_{\pm0.0}$} \\
        ~~$\bullet~$Tent~\cite{wang2021tent} & \textbf{94.5$_{\pm0.0}$} & 61.1$_{\pm0.1}$ & 93.9$_{\pm0.0}$ & 67.8$_{\pm0.7}$ & \nega1.3$_{\pm0.2}$ \\
        ~~$\bullet~$SAR~\cite{lee2023towards} & 94.1$_{\pm0.0}$ & 61.3$_{\pm0.0}$ & 97.0$_{\pm0.0}$ & 67.8$_{\pm0.7}$ & \nega0.5$_{\pm0.2}$ \\
        ~~$\bullet~$ROID~\cite{marsden2024universal} & \underline{94.4$_{\pm0.4}$} & 60.3$_{\pm0.9}$ & \underline{98.6$_{\pm0.1}$} & 68.0$_{\pm0.5}$$^*$ & \nega0.3$_{\pm0.5}$ \\
        ~~$\bullet~$DeYO~\cite{lee2024entropy} & 89.8$_{\pm1.8}$ & \textbf{63.7$_{\pm0.1}$} & 97.1$_{\pm0.4}$ & n/a$^\dagger$ & \nega0.5$_{\pm0.6}$ \\
        \rowcolor{DSBR_purple!20}~~$\bullet~$\textbf{\textit{DSBR}} (ours) & 93.9$_{\pm0.4}$ & \underline{62.6$_{\pm0.4}$} & \textbf{98.8$_{\pm0.1}$} & \underline{68.1$_{\pm0.6}$} & \textbf{\pos0.2$_{\pm0.4}$} \\
        \midrule
        ViT & 94.5$_{\pm0.0}$ & 58.9$_{\pm0.0}$ & 97.3$_{\pm0.0}$ & 57.4$_{\pm0.0}$ & \neu0.0$_{\pm0.0}$ \\
        ~~$\bullet~$Tent~\cite{wang2021tent} & 94.5$_{\pm0.0}$ & 58.9$_{\pm0.0}$ & 96.1$_{\pm0.2}$ & 57.9$_{\pm0.5}$ & \nega0.2$_{\pm0.2}$ \\
        ~~$\bullet~$SAR~\cite{lee2023towards} & 94.5$_{\pm0.0}$ & 58.9$_{\pm0.0}$ & 97.3$_{\pm0.0}$ & 57.4$_{\pm0.0}$ & \neu0.0$_{\pm0.0}$ \\
        ~~$\bullet~$ROID~\cite{marsden2024universal} & \textbf{94.6$_{\pm0.0}$} & \underline{59.1$_{\pm0.3}$} & \textbf{98.4$_{\pm0.0}$} & \textbf{59.9$_{\pm1.1}$}$^*$ & \textbf{\pos1.0$_{\pm0.4}$} \\
        ~~$\bullet~$DeYO~\cite{lee2024entropy} & \underline{94.5$_{\pm0.1}$} & 59.0$_{\pm0.1}$ & \underline{98.3$_{\pm0.0}$} & n/a$^\dagger$ & \underline{\pos0.3$_{\pm0.1}$} \\
        \rowcolor{DSBR_purple!20}~~$\bullet~$\textbf{\textit{DSBR}} (ours) & 93.0$_{\pm0.3}$ & \textbf{59.9$_{\pm0.4}$} & 97.9$_{\pm0.2}$ & \underline{58.0$_{\pm0.4}$} & \pos0.2$_{\pm0.3}$ \\
        \bottomrule
        \end{tabular}

    }
    
\end{table}

\section{ROC-AUC reveals a structural limit of EM}
\label{app:roc_auc}
\Cref{tab:results_medical_roc_auc} reports ROC-AUC scores for the same configurations.
Across the board, the gains in ROC-AUC are negligible: no method, DSBR included, exceeds $+1.0\%$ average improvement on either backbone, and several methods slightly degrade the source model's ROC-AUC despite improving its balanced accuracy substantially.
The reason is that by construction, EM sharpens the model's predictions toward the simplex corners.
Balanced accuracy depends only on the argmax and benefits from sharper peaks; ROC-AUC depends on the relative ordering of the underlying scores and is degraded when high confidence is assigned indiscriminately, including to misclassified samples.
This is a property of the entropy objective, not of any specific algorithm built on it.

\clearpage
\newpage


\end{document}